\documentclass[acmsmall,screen,nonacm]{acmart}
\AtBeginDocument{%
  }

\usepackage{xcolor}
\usepackage{makecell}
\usepackage{multirow}
\usepackage{graphicx}

\usepackage{enumitem}
\usepackage{algorithm}
\usepackage{algpseudocode}

\setcopyright{none}
\acmJournal{TSAS}
\renewcommand\footnotetextcopyrightpermission[1]{}

\begin{document}

\title{Spatial Uncertainty Analytics for Budget-Limited Inspection of Hyperspectral Scenes}

\author{Ritesh Sharma}
\affiliation{%
  \institution{Virginia Commonwealth University}
  \city{Richmond}
  \state{Virginia}
  \country{USA}}
\email{sharmar33@vcu.edu}

\author{Yuichi Motai}
\affiliation{%
  \institution{Virginia Commonwealth University}
  \city{Richmond}
  \state{Virginia}
  \country{USA}}
\email{ymotai@vcu.edu}

\renewcommand{\shortauthors}{Sharma and Motai}

\begin{abstract}
Land-cover maps classified from wide-area hyperspectral surveys are least
reliable along class boundaries and in ambiguous regions, so their errors are
spatially structured. We study how this structure can direct a limited
verification budget: when a low-altitude UAV or a ground crew can visit only a
handful of sites, where it should go depends on how classifier uncertainty is
arranged across the scene and how a fixed sensor footprint tiles that
arrangement. We estimate a per-pixel uncertainty surface from the disagreement of
an ensemble of spectral classifiers and select sites by greedy,
field-of-view-aware coverage of the uncertainty mass. On seven benchmark scenes,
the highest-uncertainty decile of pixels is $2.6$ to $7.9$ times richer in errors
than the scene as a whole, and the coverage greedy covers up to $57\%$ of the
classifier's errors with 15 sites, $1.7$ to $18$ times as many as
uncertainty-agnostic sampling of the same labelled region. A non-overlapping
selection of the footprints richest in uncertainty does equally well, so the gain
comes from scoring whole footprints by uncertainty mass. The advantage persists
under spatially blocked evaluation and with a spectral-spatial classifier,
although disagreement then ranks errors less well. It requires restricting the
surface to labelled land cover: over the whole scene, disagreement peaks on
unlabelled ground, where inspections cannot be scored. The best policy for
covering connected high-uncertainty regions depends on the scene and on the
threshold that defines them, and Gaussian-process selection covers less
uncertainty than the coverage greedy on every scene. The pipeline runs on
commodity CPUs.
\end{abstract}




\maketitle

\section{Introduction}

Scientific remote-sensing workflows increasingly use machine learning to
turn high-dimensional observations into spatially distributed maps of land
cover, vegetation state, and other environmental phenomena. A less-studied
question is how to allocate limited physical-verification effort once the map is
produced, because its reliability is \emph{spatially heterogeneous}: a model's
errors are not distributed uniformly but concentrate along class boundaries, in
mixed or transitional land cover, and in regions whose signatures the training
set under-represents. A map of where the model is \emph{uncertain} therefore has
spatial structure: it concentrates in particular fields and along their edges
rather than spreading evenly across the scene. This paper is about turning that structure into action.
When a scientific or operational workflow can afford to verify only a handful of
locations, for example with a low-altitude UAV revisiting a farm or a field crew
sampling a wetland, \emph{where} that effort is spent is a spatial-analytics
question. The answer depends on how uncertainty is arranged across the scene and
on how a fixed sensor footprint tiles that arrangement. We show that exploiting
this structure lets a fixed verification budget cover substantially more of the
model's genuine errors than uncertainty-agnostic alternatives, and we
characterize how the best strategy shifts with the scene.

We instantiate this on hyperspectral imagery. Hyperspectral sensors record
reflectance across hundreds of contiguous spectral bands, allowing materials and
vegetation states to be distinguished by their spectral signatures. This
underpins applications in precision agriculture, environmental monitoring, and
resource management, where the goal is often to locate and characterize
anomalous or ambiguous regions (stressed crops, contaminated water, unexpected
land cover) within a large scene. The follow-up verification our method plans
supports a downstream scientific decision: targeted crop or land-cover
confirmation in agriculture, ecological sampling in wetlands, and field
inspection of uncertain map regions in earth observation more broadly.

Operational remote-sensing pipelines for these applications are frequently
\emph{two-stage}. A first stage performs a wide-area airborne survey, for example
with a fixed-wing aircraft or a high-altitude UAV carrying a hyperspectral
sensor, and produces a classified map of the scene. This stage is cheap per unit
area and covers everything, but its output is only a set of \emph{model
predictions} whose reliability varies across the scene. A second stage therefore
dispatches a far more expensive resource, a low-altitude UAV for close-range
inspection or a ground crew for in-situ sampling, to verify selected locations.
This survey-then-target structure is a recurring design pattern in precision
agriculture~\cite{pretto2021flourish}, and ground-truthing is routinely used to
validate airborne hyperspectral classifications. Representative low-altitude
platforms carry lightweight hyperspectral payloads at a few tens of metres above
ground, for example $32$ to $60$\,m~\cite{guo2021maize,barbedo2022viticulture},
and their endurance is capped by battery capacity, which limits how much area a
single sortie can inspect at close range~\cite{popovic2020ipp}. The second stage
is therefore budget-limited, and the operator must decide \emph{where} to spend
scarce verification effort.

This paper addresses that second-stage decision. We assume a hyperspectral survey
has already been acquired and classified, and ask: given a spatial map of where
the classifier is most uncertain, and a UAV that can inspect only a limited number
of sites, \emph{where should it go?} This is an error-targeted inspection
prioritization problem over an already-surveyed scene, not robotic exploration of
an unknown environment or the selection of new training labels; re-surveying or ground-truthing everything is precisely what is too
expensive. The naive answer, repeatedly visiting the single most uncertain pixel,
ignores both the sensor's footprint and the fact that uncertainty concentrates
in particular parts of the scene. The right place to inspect depends not only on where
uncertainty is high but also on how it is arranged (scattered into many small
fragments or concentrated into a few large regions) and on how a fixed footprint
tiles that arrangement.

We quantify uncertainty without requiring calibrated probabilities: an ensemble
of spectral classifiers is trained on a random subset of the labelled,
dimensionality-reduced pixel spectra, and the disagreement among its members gives a normalized
uncertainty surface over the scene. We then select inspection sites greedily over
this surface under an explicit field-of-view model and compare three scores: a
peak-uncertainty policy that targets the most ambiguous remaining pixel, a
coverage policy that maximizes the mean uncertainty of a footprint's unvisited
pixels, and a dispersion policy that weights that mean by the footprint's
unvisited fraction. The dispersion score equals the uncertainty mass a footprint
newly covers divided by the constant footprint area, so it is exactly the classical
greedy algorithm for our coverage objective and inherits its $(1-1/e)$
guarantee; the other two are heuristics. To test whether uncertainty guidance is
doing real work, we compare against uncertainty-agnostic baselines (a systematic
lawnmower sweep and uniform random sampling), both over the whole scene and
restricted to the labelled region where errors are counted. To test whether the
greedy rule itself matters, we also compare against two simpler uses of the same
surface: non-maximum suppression over footprint uncertainty mass, and stratified
sampling that places one high-mass footprint in each part of the scene.

We evaluate on seven hyperspectral benchmarks. Four are standard archival
airborne scenes spanning agricultural, wetland, and urban land cover (Indian
Pines, Salinas, Kennedy Space Center, and Pavia University); three are UAV-borne
scenes at centimetre-to-decimetre resolution (WHU-Hi LongKou, HanChuan, and
HongHu~\cite{zhong2020whuhi}). We validate against ground-truth error and report
spatial metrics: uncertainty covered, coverage of high-uncertainty pixels,
coverage of \emph{connected} high-uncertainty components, and the spatial
dispersion of the chosen sites. This lets us characterize the trade-off between
depth (concentrating the budget where uncertainty mass is densest) and breadth
(reaching as many separate high-uncertainty regions as possible). We make three contributions:
\begin{itemize}[noitemsep,topsep=2pt]
\item A formulation of post-classification inspection as maximum coverage of
footprint uncertainty mass over an ensemble-disagreement uncertainty surface,
implemented in a CPU-only pipeline. The greedy algorithm we use for it is
classical; the contribution is the formulation and the evidence that footprint
uncertainty mass is a useful objective for spatial verification.
\item A ground-truth-anchored comparison against uncertainty-agnostic baselines,
scene-wide and restricted to the labelled region, and against simpler samplings
of the same surface. Scoring whole footprints by uncertainty mass, with the
coverage greedy or a non-overlapping top-mass selection, which perform alike,
covers $1.7$ to $18$ times as many errors as the best region-matched baseline
when the surface is restricted to labelled land cover.
\item An empirical characterization of the depth-versus-breadth trade-off: the
best policy for covering uncertainty is the same on every scene, but the best
policy for covering connected high-uncertainty regions depends on the scene, and,
in an exploratory analysis over seven scenes, that dependence is not predicted by
a single scalar measure of uncertainty concentration.
\end{itemize}

Section~\ref{sec:related} reviews related work. Section~\ref{sec:method} presents
the framework, Section~\ref{sec:experiments} the experimental design, and
Section~\ref{sec:results} the results. Section~\ref{sec:limitations} discusses
implications and limitations, and Section~\ref{sec:conclusion} concludes.

\section{Related Work}
\label{sec:related}

\subsection{Spatial Analytics for Scientific Decision-Making}
\label{sec:rw_spatial}

A growing body of work treats spatial structure (dependency, heterogeneity, and
autocorrelation) as the central object of analysis rather than a nuisance, using
it to drive decisions over geospatial data. Within this journal, representative
examples include modeling the spatio-temporal evolution of remote-sensing imagery
for prediction~\cite{das2020sardine}, estimating environmental fields such as air
pollution from spatial covariates~\cite{steininger2020maplur}, interpolating
sparsely sampled sensing data across space~\cite{middya2021interpolation}, and
building tooling that brings deep learning to geospatial raster
data~\cite{stewart2025torchgeo}. Our work is in this spirit but addresses a
decision that follows classification rather than classification itself: given a
spatial surface of \emph{classifier uncertainty} and a budget to physically
verify only a few locations, we ask where verification should be spent. The
analytics we exploit are the spatial arrangement of uncertainty, whether it is
scattered or concentrated, and how a fixed sensor footprint tiles that
arrangement.

\subsection{Uncertainty-Guided Sampling and Verification}
\label{sec:rw_sampling}

\paragraph{Uncertainty and active sampling in remote sensing.}
A large body of work in remote sensing studies how to select the most informative
pixels to label, under the banner of active learning. Tuia et
al.~\cite{tuia2011survey} survey the main families of query heuristics for
supervised remote sensing classification (committee or disagreement based, large
margin, and posterior-probability based), in which a heuristic ranks unlabeled
pixels by an uncertainty function and queries the most uncertain ones. More recent
surveys track the field's shift toward deep active
learning~\cite{ren2021deepactive} and benchmark uncertainty-,
representativeness-, and performance-based query strategies for hyperspectral
classification~\cite{thoreau2022active}, including Bayesian convolutional
approaches~\cite{haut2018bayesianal}. Our ensemble-disagreement measure belongs to
the committee (query-by-committee) family. The difference is the objective:
active learning selects pixels to \emph{label and improve a classifier}, whereas
we use the uncertainty surface to plan a \emph{physical inspection} for a
budget-limited UAV, where the sensor footprint and the spatial arrangement of
sites matter. A related line detects spectral anomalies directly from the
imagery, for example with background-statistics
detectors~\cite{matteoli2010anomaly} and a broad family of hyperspectral anomaly
detectors~\cite{su2022anomaly}; we instead treat classifier uncertainty as the
cue for where physical inspection is most needed.

\paragraph{Ensemble uncertainty.}
We obtain per-pixel uncertainty from the disagreement of an ensemble of tree
classifiers. Random forests~\cite{breiman2001random} are a standard, robust
ensemble for hyperspectral pixel classification~\cite{ham2005randomforest}, and
disagreement across members is a simple, model-agnostic uncertainty proxy that
requires no calibrated probabilities; in active learning, ensemble disagreement
has outperformed Monte-Carlo dropout as an uncertainty
estimate~\cite{beluch2018ensembles}.

\paragraph{Hyperspectral classification.}
The first stage of our pipeline is an ordinary hyperspectral pixel classifier, a
mature area: classical pipelines pair dimensionality reduction with
support-vector~\cite{melgani2004svm} or random-forest classifiers and exploit
spatial-spectral structure~\cite{fauvel2013spatialspectral,ghamisi2017advances},
while recent work is dominated by deep spectral-spatial
networks~\cite{li2019deephsi}, including transformer
architectures~\cite{hong2022spectralformer} and methods for the
few-labeled-sample regime~\cite{jia2021fewlabeled}. Our contribution is
orthogonal: we consume \emph{any} classifier's predictions and their
disagreement, and treat the classifier as a modular component
(Section~\ref{sec:estimator}).

\paragraph{Modern uncertainty quantification.}
Beyond tree-ensemble disagreement, a substantial literature estimates predictive
uncertainty in deep models: deep
ensembles~\cite{lakshminarayanan2017deepensembles} average independently trained
networks, Monte-Carlo dropout~\cite{gal2016dropout} interprets dropout at
inference as approximate Bayesian sampling, evidential
models~\cite{sensoy2018evidential} predict a distribution over class
probabilities, and Kendall and Gal~\cite{kendall2017uncertainties} distinguish
aleatoric from epistemic uncertainty in vision; surveys catalogue these
techniques and their open challenges~\cite{abdar2021uqreview,gawlikowski2023uqsurvey}.
Such estimates have been used in hyperspectral classification, for example to
drive Bayesian active learning~\cite{haut2018bayesianal}, and in pixel-wise
remote-sensing segmentation~\cite{haas2021sar} to flag unreliable predictions.
Our contribution is the inspection-planning layer over an uncertainty surface,
not the estimator; we adopt a CPU-friendly tree ensemble and show in
Section~\ref{sec:estimator} that the planner works unchanged with deep
estimators.

\subsection{Informative Path Planning and Spatial Coverage}
\label{sec:rw_ipp}

\paragraph{Informative path planning.}
Deciding where a mobile sensor should take measurements under a budget is the
informative path planning (IPP) problem; UAV path and coverage planning more
broadly is a large field, spanning general path
planning~\cite{jones2023pathsurvey} and systematic coverage path
planning~\cite{galceran2013coverage,cabreira2019coverage}. Singh et al.~\cite{singh2009informative}
give approximation algorithms for single- and multi-robot informative sensing,
modeling the phenomenon with a Gaussian process and quantifying information by
mutual information between visited and unvisited locations. Popovi\'c et
al.~\cite{popovic2020ipp} present an IPP framework for UAV-based terrain
monitoring that contrasts adaptive planning against passive lawnmower coverage,
the same baseline we adopt. More recent work plans \emph{online} to reduce model
uncertainty: R\"uckin et al.~\cite{ruckin2023ipp} guide a UAV to collect
uncertain images for active-learning retraining of a segmentation model,
and Krestenitis et al.~\cite{krestenitis2024overfomo} modulate UAV scanning speed
by detection confidence. These methods adapt a trajectory in an initially
\emph{unknown} scene. Closest to our setting, Xu~\cite{xu2026energy} flies a
coarse survey, thresholds an MC-dropout entropy map and routes a low-altitude
re-inspection of every flagged cell, evaluated in simulation on synthetic
hotspots; we instead select a fixed number of sites by footprint coverage of
uncertainty mass and validate them against ground-truth classifier errors. More
generally, our setting is narrower and complementary: we prioritize \emph{physical verification} over an
\emph{already-surveyed} scene, selecting sites over a \emph{fixed}, precomputed
classifier-uncertainty surface. Cost-benefit greedy planners such as
GCB~\cite{zhang2016submodular} maximize a submodular reward per unit travel with
approximation guarantees, and submodular-orienteering formulations jointly
optimize information and tour length. Our travel-aware variant
(Section~\ref{sec:costaware}) is a lightweight heuristic in this family; a
treatment with routing guarantees and a flight-cost model is outside the scope of
this paper.

\paragraph{Coverage and orienteering formulations.}
The objective we optimize, the uncertainty mass falling within a fixed number of
field-of-view footprints, is maximum coverage under a cardinality constraint, the
unit-cost case of budgeted maximum coverage~\cite{khuller1999budgeted}. Its
tour-constrained counterpart, in which the sites must form a length-bounded
route, is submodular orienteering. For the cardinality-constrained problem the
greedy algorithm that adds the site of largest marginal gain achieves a
$(1-1/e)$ approximation~\cite{nemhauser1978submodular}, and informative-sensing
formulations exploit the same structure~\cite{singh2009informative}. Our
dispersion policy is exactly this greedy algorithm on the candidate grid
(Section~\ref{sec:policies}); the peak and greedy-coverage policies are
heuristics. We do not solve the orienteering problem; Section~\ref{sec:cost}
reports the travel that each policy's sites imply.

\subsection{Positioning of This Work}
\label{sec:rw_positioning}

Our contribution sits at the intersection of these threads. Unlike active
learning, we do not select pixels to \emph{improve} a classifier but to
\emph{physically verify} an existing one, so the sensor footprint and inter-site
geometry, irrelevant when merely querying labels, become central. Unlike online
IPP, we plan over a \emph{fixed}, already-computed uncertainty surface, which
matches the operational reality that re-surveying is the expensive step we are
trying to avoid. Unlike the information-theoretic IPP tradition, whose objective
is mutual information about a latent field, we optimize direct \emph{coverage of
uncertainty mass} under a footprint model, which we show is far better matched
to physical inspection (Section~\ref{sec:gpmi}). And unlike work that stops at
aggregate uncertainty metrics, we validate the pipeline against ground-truth
error (Sections~\ref{sec:error_corr}--\ref{sec:error_reduction}) and characterize
how the \emph{spatial scale} of uncertainty determines which strategy wins. The
result is not a new uncertainty estimator or a new solver, but a
spatial-analytics treatment of a decision problem, where to verify, that prior
work has largely left implicit.

\section{Spatial Uncertainty-Guided Inspection Framework}
\label{sec:method}

Our pipeline has two stages that mirror the operational workflow: we estimate a
spatial surface of classifier uncertainty over an already-classified scene
(Section~\ref{sec:method_unc}), then select a budget-limited set of inspection
sites that cover as much of that uncertainty as possible under a sensor-footprint
model (Section~\ref{sec:method_plan}). The stages are decoupled: the planner
consumes only a normalized scalar surface and is agnostic to how it was produced,
a modularity we exploit in Section~\ref{sec:estimator}.

\subsection{Problem Formulation}
\label{sec:problem}
We are given a classified hyperspectral scene and a per-pixel uncertainty surface
$U : \Omega \to [0,1]$ over the pixel grid $\Omega$. A sensor with a square field
of view (FOV) of radius $r$ inspects, from each chosen site, the footprint of
pixels within that radius. Given a budget of $K$ sites, we seek a set
$S = \{s_1, \dots, s_K\}$ of footprint centres whose union of footprints
$\mathrm{cov}(S) = \bigcup_{s \in S} \mathrm{FOV}_r(s)$ maximizes the covered
uncertainty mass,
\begin{equation}
\label{eq:objective}
\max_{|S| = K} \; \sum_{p \in \mathrm{cov}(S)} U(p).
\end{equation}
The objective is monotone and submodular in $S$: adding a site never decreases
the covered mass, and its marginal gain shrinks as coverage grows.
Problem~\eqref{eq:objective} is maximum coverage under a cardinality constraint,
the unit-cost case of budgeted maximum coverage~\cite{khuller1999budgeted}, and
the greedy algorithm that at each step adds the site with the largest
\emph{marginal gain}, the uncertainty mass it newly covers, attains a $(1-1/e)$
approximation of the optimum over the candidate sites~\cite{nemhauser1978submodular}.
The same structure underlies near-optimal sensor placement~\cite{krause2008near},
which our Gaussian-process baseline (Section~\ref{sec:gpmi}) instantiates. The
guarantee applies only to greedy selection on the exact marginal gain. Among our
policies, dispersion (Section~\ref{sec:policies}) is exactly this greedy
algorithm; the peak and greedy-coverage policies use other scores and carry no
guarantee, and the travel-aware variant (Section~\ref{sec:costaware}) trades
covered mass against travel and forgoes it as well.

\subsection{Constructing the Spatial Uncertainty Surface}
\label{sec:method_unc}

Let the hyperspectral scene be a cube $X \in \mathbb{R}^{H \times W \times B}$
with $B$ spectral bands. We reshape $X$ into $HW$ pixel spectra and reduce each to
$d$ principal components via PCA, followed by standardization, giving a feature
matrix $Z \in \mathbb{R}^{HW \times d}$. We train an ensemble of $M$ classifiers
$\{f_1, \dots, f_M\}$ that differ in model family and random state. Each member
is trained on its own, independently drawn random fraction $\tau$ of the labelled
pixels (those with a ground-truth class), so the committee varies both the model
and the training sample, as in bagging. Each member then predicts a class label for every
pixel, and we define the pixel-level uncertainty as the mean \emph{pairwise
disagreement} among members,
\begin{equation}
\label{eq:unc}
u(p) = \frac{2}{M(M-1)} \sum_{1 \le i < j \le M}
       \mathbb{1}\!\left[\, f_i(z_p) \neq f_j(z_p) \,\right],
\end{equation}
the fraction of member pairs that predict different classes. Unlike the variance
of integer class indices, this measure is invariant to how classes are numbered.
With $M=5$ it takes only a handful of distinct values (at most seven in our
scenes), and most pixels have $u=0$. We min--max normalize $u$ to $[0,1]$ over the
\emph{labelled} region to obtain $U$; pixels with no ground-truth label are
assigned zero uncertainty and are not selected by the uncertainty-driven
policies. Unlabelled pixels are not all out of distribution: they mix unlabelled
parts of the trained classes, mixed pixels at field and road edges, and land cover
the benchmark never labelled. Without ground truth, disagreement there cannot be
scored as error and cannot be split into in-class ambiguity and novelty
(Section~\ref{sec:roi_control} removes the mask and shows why it matters). Because
$53$ to $95\%$ of labelled pixels have $u=0$, the minimum is zero on every
surface and the normalization is a rescaling by the largest labelled
disagreement, which is $1$ (all five members predict different classes) on most
surfaces and $0.9$ on Pavia University and two LongKou seeds. A rescaling leaves
the sites chosen by every policy except the cost-aware one unchanged. Since $u$ is
a multiple of $0.1$, it also leaves the
pixels above every threshold we use for connected anomalies
(Section~\ref{sec:metrics}) unchanged: $U \ge 0.7$ selects exactly the pixels with
$u \ge 0.7$, those on which the five members predict at least three different
classes.
The delivered map is the members' majority vote.

\paragraph{Design rationale.}
First, we use \emph{committee disagreement} rather than a single model's
predictive entropy because disagreement among members trained on different
samples of scarce labels is a proxy for epistemic uncertainty, due to limited or
unrepresentative training data, which is the uncertainty a physical re-inspection
can resolve. Resampling matters: members that share one training subset rank
errors markedly worse (Section~\ref{sec:tau}). Second, we adopt a
\emph{lightweight tree ensemble} rather than a deep Bayesian or MC-dropout
estimator: the planner consumes only a scalar surface, so the estimator can be
upgraded without altering it, and Section~\ref{sec:estimator} shows that the
tree ensemble ranks errors at least as well as three neural alternatives on
Indian Pines. Third, we compute disagreement over per-member \emph{class predictions}
rather than averaged probabilities because this requires no probability
calibration across heterogeneous model families.

\subsection{Candidate Footprints and Greedy Planning}
\label{sec:method_plan}

Exactly solving~\eqref{eq:objective} is NP-hard, so we proceed greedily: at each
of the $K$ steps the planner scores every candidate site, selects the best, and
marks its footprint visited. The three greedy uncertainty-driven policies differ only in
the score; the baselines ignore $U$. Candidate sites for the greedy-coverage and
dispersion policies are the centres of square FOV windows on a regular grid of
stride $s$, placed so that every candidate footprint lies inside the scene.

\subsection{Uncertainty-Guided Inspection Policies}
\label{sec:policies}

\subsubsection{Peak-Uncertainty Selection}
Select the unvisited pixel of maximum uncertainty and inspect the footprint
centred on it. This maximizes point uncertainty and ignores that the footprint
covers a neighbourhood. Because $u$ takes only a few distinct values, many pixels
tie for the maximum; ties go to the first pixel in scan order.
Section~\ref{sec:tau} repeats the analysis with ties broken uniformly at random.

\subsubsection{Greedy Uncertainty Coverage}
Let $F$ be a candidate footprint, $V$ the set of already-visited pixels, and
$U_F = \{p \in F : p \notin V\}$ its unvisited pixels. The greedy-coverage score
is the mean uncertainty over the unvisited portion of the footprint,
\begin{equation}
\label{eq:greedy}
s_{\mathrm{cov}}(F) = \frac{1}{|U_F|} \sum_{p \in U_F} U(p),
\end{equation}
and the planner inspects $\hat F = \arg\max_F s_{\mathrm{cov}}(F)$, with the
waypoint at the footprint centre. This score is \emph{not} the marginal gain: a
footprint with a few unvisited high-uncertainty pixels left at its edge can
outscore a fresh footprint that would cover far more uncertainty mass, so the
policy tends to return to partly inspected areas.

\subsubsection{Uncertainty-Mass Coverage (Dispersion)}
On the same candidate grid, the dispersion score weights the mean unvisited
uncertainty by the fraction of the footprint still unvisited,
\begin{equation}
\label{eq:disp}
s_{\mathrm{disp}}(F) = \left(\frac{1}{|U_F|} \sum_{p \in U_F} U(p)\right)
                       \cdot \frac{|U_F|}{|F|}
                     = \frac{1}{|F|}\sum_{p \in U_F} U(p),
\end{equation}
which down-weights footprints that overlap inspected area. Since every candidate
footprint lies inside the scene, $|F| = (2r+1)^2$ is the same for all candidates,
so maximizing $s_{\mathrm{disp}}$ is the same as maximizing the uncertainty mass
the footprint newly covers, the marginal gain of~\eqref{eq:objective}. The
dispersion policy is therefore the classical greedy algorithm for maximum
coverage and inherits its $(1-1/e)$ guarantee relative to the best $K$ grid
footprints. We keep the name because the overlap penalty is what spreads its
sites across the scene. We use a candidate-grid stride of $s = 5$ pixels for the
greedy-coverage and dispersion policies.

\subsubsection{Cost-Aware Dispersion}
Objective~\eqref{eq:objective} counts sites equally, but a UAV pays a travel cost
between them. We therefore add a distance penalty: when the last selected site is
$s_{\mathrm{prev}}$, a candidate footprint $F$ centred at $c$ is scored
\begin{equation}
\label{eq:costaware}
s_{\lambda}(F) = s_{\mathrm{disp}}(F) \;-\; \lambda \,
                 \frac{\lVert c - s_{\mathrm{prev}} \rVert_2}{\mathrm{diag}(\Omega)},
\end{equation}
where $\mathrm{diag}(\Omega)$ is the scene diagonal (making $\lambda$
scene-independent) and $\lambda \ge 0$ trades covered mass against travel
distance. At $\lambda = 0$ this is the dispersion policy; for $\lambda > 0$ it is
a travel-aware heuristic, not an orienteering solver. Section~\ref{sec:costaware}
analyzes the trade-off.

\subsection{Baseline Inspection Strategies}
\label{sec:baselines}

\textit{Lawnmower.} A boustrophedon sweep over a non-overlapping grid of stride
$2r+1$ (one footprint), starting at a corner of the scene and truncated after $K$
footprints.

\textit{Random.} Uniformly sampled unvisited pixels, each inspected with a
footprint centred on it. Every random baseline is averaged over $20$ independent
draws per classifier seed.

These baselines ignore uncertainty, so any advantage of the uncertainty-driven
policies over them reflects the value of steering inspection toward model
ambiguity. Section~\ref{sec:protocol} adds versions restricted to the labelled
region.

Two further baselines do use the surface and test whether the marginal-gain
greedy adds anything over simpler spatial sampling of it.
\textit{Uncertainty-ranked non-maximum suppression (NMS)} ranks the candidate
footprints once by the uncertainty mass they hold and takes them in that order,
skipping any footprint that overlaps one already taken. It never revisits ground,
but unlike dispersion it cannot take a footprint that partly overlaps an
inspected one even when that footprint still holds the most new mass.
\textit{Stratified uncertainty sampling} partitions the labelled region into $K$
compact strata ($k$-means on pixel coordinates) and takes, in each stratum, the
candidate footprint with the most uncertainty mass, so it guarantees one site per
part of the scene. Both use the same candidate grid and footprint as dispersion,
and the result tables list them with the uncertainty-driven policies.

\subsection{Computational Complexity}
\label{sec:complexity}
Algorithm~\ref{alg:planner} states the greedy planner shared by the
uncertainty-driven policies; they differ only in the score $s(\cdot)$ at line~4.
Let $|C|$ be the number of candidate footprints on the stride-$s$ grid. Each of
the $K$ rounds scores every candidate, and scoring reads an $O(r^2)$ footprint
(or $O(1)$ with an integral image), so the planner runs in $O(K\,|C|\,r^2)$ time;
since $|C| = O(HW/s^2)$, this is linear in the number of pixels for fixed $r$,
$s$ and $K$. Gaussian-process informative planning, by contrast, costs $O(|C|^2)$
per step for mutual-information selection and $O(|C|^3)$ for the kernel inverse
(Section~\ref{sec:gpmi}), so the coverage planner scales to the largest scenes we
study on CPU. The $(1-1/e)$ guarantee depends only on the score: it holds for
dispersion, whose score is the exact marginal gain, and not for the other
policies, although all share the same running time.

\begin{algorithm}[t]
\caption{Greedy field-of-view-aware inspection planning.}
\label{alg:planner}
\begin{algorithmic}[1]
\Require uncertainty surface $U$; FOV radius $r$; budget $K$; candidate grid
  $C$; score $s(\cdot)$
\State $V \gets \varnothing$ \Comment{visited pixel set}
\State $S \gets \langle\,\rangle$ \Comment{ordered inspection sites}
\For{$k = 1$ \textbf{to} $K$}
  \State $c^\star \gets \arg\max_{c \in C}\; s\!\left(F_r(c) \setminus V\right)$
     \Comment{best-scoring footprint over unvisited pixels}
  \State append $c^\star$ to $S$
  \State $V \gets V \cup F_r(c^\star)$ \Comment{mark the executed footprint visited}
\EndFor
\State \Return $S$, $V$
\end{algorithmic}
\end{algorithm}

\section{Experimental Design}
\label{sec:experiments}

\subsection{Research Questions}
\label{sec:rqs}

The premise that inspecting uncertain regions is worthwhile rests on the spatial
arrangement of classifier uncertainty being a faithful proxy for the arrangement
of classifier \emph{error}, so we first ask whether that correspondence holds.
We then ask whether steering inspection by uncertainty covers more genuine
misclassifications than uncertainty-agnostic strategies under a fixed number of
sites, including strategies that know where the labelled region is and
strategies that sample the uncertainty surface without the marginal-gain rule,
and whether the answer survives a spatially blocked evaluation. Because the
uncertainty-driven policies differ chiefly in \emph{how} they distribute effort,
we next examine how the spatial structure of uncertainty shapes their relative
performance, in particular the trade-off between depth and breadth, and whether
that outcome is predictable from the geometry of the surface alone. We then test
robustness to the training fraction, footprint, budget, ensemble,
connected-component threshold, uncertainty estimator and classifier features. Finally, since sites
count equally only in the abstract, we ask what travel the selected sites imply
and how a travel penalty changes the picture.

\subsection{Datasets and Spatial Characteristics}
\label{sec:datasets}

We evaluate on four archival airborne scenes spanning different land cover:
Indian Pines ($145 \times 145$, 200 bands, 16 classes; agricultural), Salinas
($512 \times 217$, 204 bands, 16 classes; agricultural, with larger and more
contiguous fields), Kennedy Space Center (KSC, $512 \times 614$, 176 bands, 13
classes; coastal wetland), and Pavia University ($610 \times 340$, 103 bands, 9
classes; urban). These are standard benchmarks with ground-truth labels and
relatively coarse resolution (Indian Pines is AVIRIS data at about
$20$\,m/pixel), used as controlled proxies for the second-stage problem.

To evaluate at the resolution the method targets, we also use the WHU-Hi
benchmark~\cite{zhong2020whuhi}: three UAV-borne hyperspectral scenes acquired
with a Headwall Nano-Hyperspec sensor over farmland in Hubei, China:
\emph{LongKou} ($550\times400$, 270 bands, $\sim$0.46\,m/pixel; 9 classes),
\emph{HanChuan} ($1217\times303$, 274 bands, $\sim$0.11\,m/pixel; 16 classes, a
rural-urban fringe), and \emph{HongHu} ($940\times475$, 270 bands,
$\sim$0.04\,m/pixel; 22 classes). These are genuine low-altitude UAV surveys.

In all seven benchmarks only part of the scene carries a ground-truth label: the
labelled share is $49\%$ on Indian Pines and Salinas, $2\%$ on KSC, $21\%$ on
Pavia University, and $93$, $70$ and $87\%$ on LongKou, HanChuan and HongHu. This
sparsity is a property of the benchmark labels, not of the scenes, and it matters
for the baselines (Section~\ref{sec:protocol}).

\subsection{Experimental Protocol}
\label{sec:protocol}

We reduce each spectrum to $30$ principal components and train an ensemble of
$M=5$ classifiers, a mix of random forests ($50$ trees, max depth $15$) and
gradient-boosted trees ($50$ estimators, max depth $5$) with different random
states, each trained on its own random fraction $\tau = 0.1$ of the labelled
pixels. PCA and standardization are computed in
double precision: in single precision the trailing components are dominated by
rounding error, which standardization amplifies, and member predictions then vary
between machines. With this choice every surface is reproducible exactly; all were
computed with scikit-learn 1.9.0, NumPy 2.4.6 and SciPy 1.17.1. The FOV radius is $r=7$ pixels (a $15\times15$ footprint) and the budget
is $K=15$ sites. We fix the footprint in \emph{pixels}, so its physical size
varies with each scene's ground sampling distance (GSD), from $300\times300$\,m
on Indian Pines to $0.6\times0.6$\,m on HongHu (Table~\ref{tab:physfov}). This
holds the combinatorial planning problem fixed across scenes, so the comparison
between policies is controlled, but it means absolute coverage is not physically
comparable across datasets, and on the archival scenes a footprint of a few
hundred metres corresponds to a flight altitude of roughly $200$\,m with a
typical $70^\circ$ camera. The archival scenes therefore test the spatial
analytics rather than a specific platform. We repeat each experiment over
multiple classifier seeds ($10$ for Indian Pines, $5$ for every other scene) and
report mean $\pm$ standard deviation over seeds. Because every policy is
evaluated on the same surfaces, we compare policies by paired differences over
seeds and report the mean difference with a $95\%$ Student-$t$ interval and the
number of seeds on which one policy beats the other. With five seeds these
intervals are wide; they describe the seeds we ran, not a population of scenes.
All experiments run on CPU: on 16 cores,
building one seed's surface takes about 20 minutes on the largest scene (HongHu),
and a 15-site dispersion plan takes $0.004$ to $0.09$\,s on one core
(Section~\ref{sec:complexity}); stratified sampling, whose $k$-means step
dominates, takes $0.6$ to $2.6$\,s.

Sites, footprints and distances live in raster coordinates. A footprint is a
square block of pixels and a route length is a pixel distance scaled by the GSD;
we do not model terrain, camera tilt, georeferencing error or the mapping from a
waypoint to a ground footprint, which a deployment would add between the planner
and the platform.

\begin{table}[t]
  \caption{Physical size of the fixed $15\times15$-pixel sensor footprint across
  scenes, given each scene's ground sampling distance (GSD). Holding the
  footprint fixed in pixels controls the planning problem for the cross-scene
  comparison; absolute coverage is therefore not physically comparable across
  datasets.}
   \vspace{-0.3cm}
  \label{tab:physfov}
  \begin{tabular}{lccc}
    \toprule
    Scene & GSD (m/px) & Footprint (m) & Area (m$^2$) \\
    \midrule
    Indian Pines & $20.0$  & $300 \times 300$ & $90{,}000$ \\
    Salinas      & $3.7$   & $55.5 \times 55.5$ & $3{,}080$ \\
    KSC          & $18.0$  & $270 \times 270$ & $72{,}900$ \\
    Pavia U.     & $1.3$   & $19.5 \times 19.5$ & $380$ \\
    LongKou      & $0.46$  & $6.9 \times 6.9$ & $48$ \\
    HanChuan     & $0.11$  & $1.6 \times 1.6$ & $3$ \\
    HongHu       & $0.04$  & $0.6 \times 0.6$ & $0.4$ \\
    \bottomrule
  \end{tabular}
\end{table}

\paragraph{Selection region and region-matched baselines.}
The uncertainty surface is defined on the labelled pixels, so the
uncertainty-driven policies select within the labelled region, which is also
where classifier errors are counted. The scene-wide lawnmower and random
baselines sweep the whole scene, as an operator without an uncertainty map would
in a fully classified scene. In the benchmarks, however, half of Salinas and
most of KSC and Pavia University is unlabelled, so a scene-wide baseline spends
much of its budget where no error can be counted. To separate the value of
the uncertainty surface from knowledge of where the labels are, we also report
three baselines restricted to the labelled region: \emph{random (labelled)},
which samples labelled pixels uniformly; \emph{lawnmower (labelled)}, which runs
the boustrophedon sweep over only those grid cells whose footprint contains a
labelled pixel; and \emph{systematic (labelled)}, which takes $K$ of those cells
evenly spaced along the sweep, a systematic spatial sample of the labelled area.
These know where the labels are but not where the classifier is uncertain. They
use the same footprints, error definitions and classifier seeds as the other
policies.

\subsection{Evaluation Metrics}
\label{sec:metrics}

Our primary, ground-truth-anchored metric is \emph{error coverage}: the fraction
of the classifier's misclassified labelled pixels (majority vote against ground
truth) that fall within an inspected footprint. \emph{Inspection precision} is the
fraction of the labelled pixels an inspection visits that are errors. We also
report four metrics on the uncertainty surface itself: (1) \emph{uncertainty
covered}, the share of the scene's total uncertainty mass $\sum_p U(p)$ that lies
inside the inspected footprints, which is objective~\eqref{eq:objective}
normalized by its largest possible value; it measures how much of the surface an
inspection reaches, not any change in the classifier, which is not retrained;
(2) \emph{connected-anomaly coverage}, the
mean, over connected components of $\{U \ge 0.7\}$ (4-connectivity), of the
fraction of each component's pixels inside an inspected footprint, set to $1$ if
no pixel exceeds the threshold. Because this mean weights a two-pixel fragment
like a large region, we also report two variants: the size-weighted coverage,
the share of all pixels in such components that are inspected, and the share of
components that an inspection touches at all; (3) \emph{high-uncertainty coverage}, the
fraction of labelled pixels with $U>0$ that are visited (the median of $U$ is
zero on every scene, so these are the above-median pixels, those on which at
least two members disagree); and (4)
\emph{waypoint dispersion}, the mean pairwise distance between sites, which is
descriptive. The threshold of $0.7$ selects the clearest high-uncertainty
structures; Section~\ref{sec:tau} sweeps it.

\section{Results and Spatial Analysis}
\label{sec:results}

\subsection{Validating Uncertainty as a Proxy for Classification Error}
\label{sec:error_corr}

Our metrics are derived from the same ensemble that produces the uncertainty
surface, so they could be self-referential. To test whether disagreement
corresponds to genuine \emph{misclassification}, we use the ground-truth labels:
on the labelled pixels of each scene we take the ensemble's majority-vote
prediction, mark the pixels it gets wrong, and measure how well $u(p)$ ranks
those errors by the area under the ROC curve (AUC). Table~\ref{tab:auc} reports
the result. In this protocol each member has seen a tenth of the evaluated pixels,
and about $40\%$ of them have been seen by at least one member, which inflates the
AUC. We therefore also report a leakage-free variant and treat it as the primary
measure: we split the labelled pixels once into a training pool ($40\%$) and a
held-out test set ($60\%$), draw every member's $\tau$-fraction only from the
pool, and evaluate AUC only on held-out pixels. The held-out AUC is
$0.035$ to $0.078$ below the main AUC but lies between $0.74$ and $0.82$ on every
scene, so disagreement ranks errors on pixels no member has seen.

Held-out pixels can still be immediate neighbours of training pixels, and
neighbouring pixels in a hyperspectral scene are strongly correlated, so a
pixel-wise split can overstate how well the surface transfers to ground the
classifier has not seen. A third, spatially blocked protocol removes this: we
tile the scene into $32\times32$-pixel blocks, assign whole blocks at random to
the training pool until it holds $40\%$ of the labelled pixels, and test on the
labelled pixels of the remaining blocks that lie more than $7$ pixels (one
footprint radius) from any pool block, so no test pixel borders a training pixel
and no footprint centred on a test pixel reaches into a pool block. The blocked
AUC (Table~\ref{tab:auc}) stays close to the held-out AUC on the UAV-borne scenes
($0.740$ to $0.792$) but falls to $0.630$ to $0.692$ on the archival scenes, with
seed spreads of up to $0.08$ because a few blocks decide which fields the
classifier has seen. Disagreement still ranks errors above chance everywhere, but
on the archival scenes it ranks them noticeably worse on ground far from any
training pixel. We also plan inside the test blocks alone and score error
coverage there (Section~\ref{sec:blocked_eval}).

\begin{table}[t]
  \caption{Does disagreement predict error? ROC-AUC of the uncertainty score
  $u(p)$ for ranking misclassified labelled pixels (majority-vote error against
  ground truth), mean $\pm$ std over seeds (Indian Pines: 10, others: 5). ``AUC
  (held-out)'' repeats the test with evaluation pixels excluded from every
  member's training set (Section~\ref{sec:error_corr}) and is the primary
  measure; ``AUC (blocked)'' trains on whole $32\times32$ blocks and tests on
  labelled pixels more than $7$ pixels from any of them. ``AP, held-out'' is the
  average precision of $u$ on held-out pixels, next to the held-out error
  prevalence, which is the AP of an uninformative score. ``Top-10\% error'' is the error rate among the
  $10\%$ of labelled pixels ranked highest by $u$, with ties at the cut broken at
  random; ``Error where $u>0$'' is the error rate among pixels on which at least
  two members disagree. Both are to be read against the scene's overall error
  rate.}
   \vspace{-0.3cm}
  \label{tab:auc}
  \footnotesize
  \setlength{\tabcolsep}{3pt}
  \begin{tabular}{lccccccc}
    \toprule
    Scene & AUC & \makecell{AUC\\(held-out)} & \makecell{AUC\\(blocked)} & \makecell{AP, held-out\\(prevalence)} & \makecell{Error\\rate} & \makecell{Top-10\%\\error} & \makecell{Error\\where $u>0$} \\
    \midrule
    Indian Pines & $0.789 \pm 0.006$ & $0.742 \pm 0.011$ & $0.630 \pm 0.070$ & $0.48$ ($0.26$) & $22.8\%$ & $59.3\%$ & $41.2\%$ \\
    Salinas      & $0.825 \pm 0.004$ & $0.747 \pm 0.007$ & $0.648 \pm 0.052$ & $0.26$ ($0.06$) & $5.8\%$ & $38.1\%$ & $37.5\%$ \\
    KSC          & $0.825 \pm 0.009$ & $0.778 \pm 0.011$ & $0.692 \pm 0.081$ & $0.51$ ($0.25$) & $22.9\%$ & $65.9\%$ & $44.8\%$ \\
    Pavia U.     & $0.827 \pm 0.004$ & $0.767 \pm 0.005$ & $0.652 \pm 0.045$ & $0.32$ ($0.10$) & $8.9\%$ & $44.1\%$ & $37.9\%$ \\
    \midrule
    LongKou      & $0.870 \pm 0.001$ & $0.817 \pm 0.006$ & $0.792 \pm 0.021$ & $0.31$ ($0.03$) & $2.2\%$ & $17.3\%$ & $35.8\%$ \\
    HanChuan     & $0.809 \pm 0.001$ & $0.760 \pm 0.002$ & $0.740 \pm 0.019$ & $0.40$ ($0.14$) & $12.6\%$ & $56.5\%$ & $50.4\%$ \\
    HongHu       & $0.784 \pm 0.001$ & $0.749 \pm 0.001$ & $0.748 \pm 0.017$ & $0.46$ ($0.19$) & $17.5\%$ & $66.1\%$ & $55.6\%$ \\
    \bottomrule
  \end{tabular}
\end{table}

Disagreement is a consistent predictor of misclassification: the held-out AUC
lies between $0.74$ and $0.82$ on every scene, well above the $0.5$ chance level,
with small variance across seeds. The top decile of pixels ranked by $u$ is $2.6$
to $7.9$ times richer in errors than the scene as a whole: on Indian Pines and KSC
$59$ and $66\%$ of it is misclassified against base error rates near $23\%$, and on
the accurately classified Salinas and LongKou it is $38.1\%$ and $17.3\%$ against
$5.8\%$ and $2.2\%$. Between $53\%$ (Indian Pines) and $95\%$ (LongKou) of the
labelled pixels have $u=0$, and among pixels where members disagree the error rate
is $36$ to $56\%$. AUC is insensitive to how rare errors are, so we also report
average precision (AP), whose chance level is the error prevalence. On held-out
pixels the AP is $0.26$ to $0.51$, $1.8$ (Indian Pines) to about $10$ (LongKou) times the prevalence; the ratio is largest where errors are rarest. Covering
high-uncertainty regions is therefore a sound proxy for covering classifier
errors.

\subsection{Error Coverage under a Fixed Inspection Budget}
\label{sec:error_reduction}

We first ask the operational question: does steering inspection by the spatial
structure of classifier uncertainty cover more genuine misclassifications than
uncertainty-agnostic strategies? For each scene we measure what fraction of the
majority-vote errors fall within an inspected footprint under a fixed budget of
15 sites. This test references ground truth that the first-stage classifier never
resolved, so it cannot be an artifact of the uncertainty measure. It measures
error \emph{coverage}, whether an inspected footprint contains a genuine error,
not error \emph{correction}: the study is an offline analysis of where
verification effort should go, and the correction itself is the operator's.

\begin{figure*}[t]
  \centering
  \includegraphics[width=\linewidth]{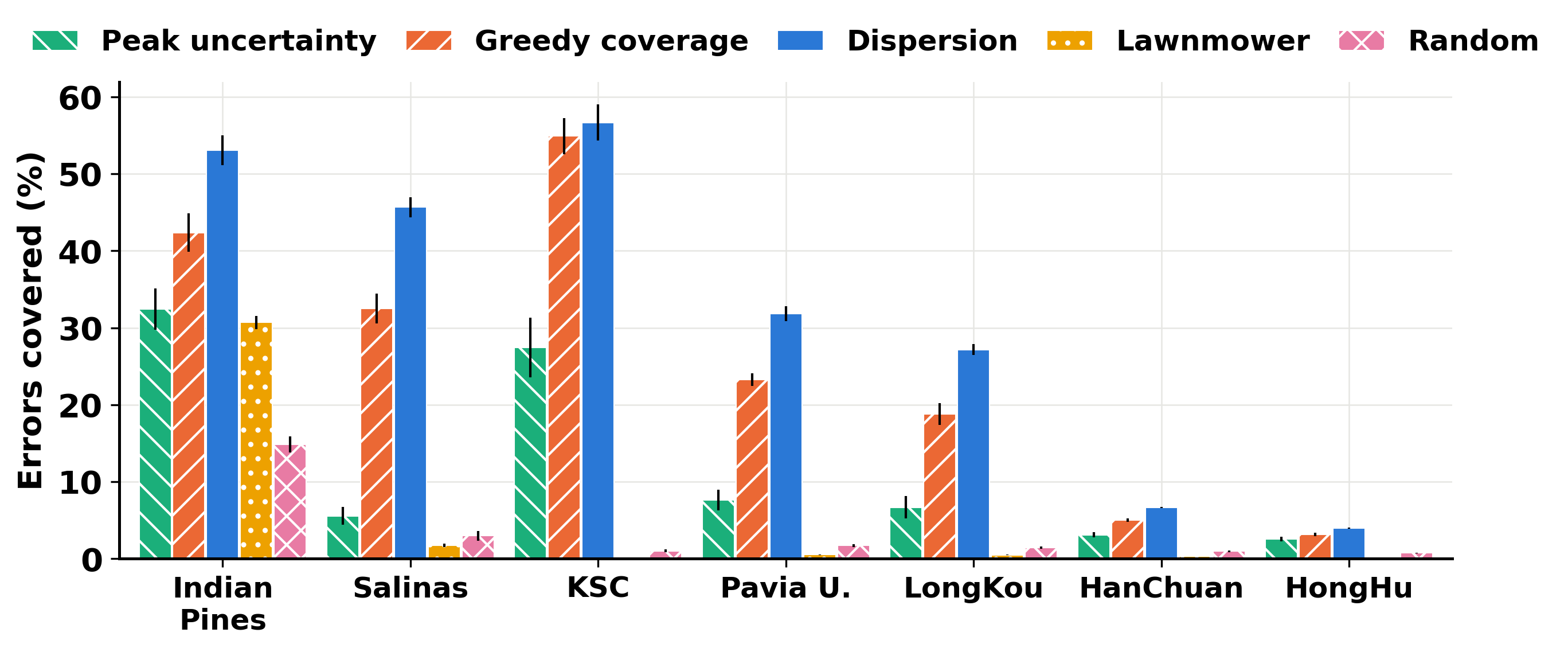}
  \vspace{-1.0cm}
  \caption{\textbf{Error coverage with a budget of 15 sites.} Fraction of genuine
  classifier errors covered by each policy across the seven scenes (mean $\pm$
  std). Lawnmower and Random are the scene-wide baselines. Dispersion covers the most errors among the plotted policies; uncertainty-ranked NMS, which is within $0.4$ points of it, stratified sampling and the region-matched baselines are in Table~\ref{tab:error_reduction}.}
  \label{fig:error_reduction}
  \Description{Grouped bar chart of the percentage of classifier errors covered by five policies on seven scenes; dispersion has the tallest bar on every scene.}
\end{figure*}

Table~\ref{tab:error_reduction} and Figure~\ref{fig:error_reduction} report the
result. The two policies that score whole footprints by the uncertainty mass they
cover, dispersion and uncertainty-ranked NMS, cover the most errors on every
scene, from $56.7\%$ on KSC and $53.0\%$ on Indian Pines to $4.0\%$ on HongHu, and they are within $0.4$ points of each other. Paired over seeds, dispersion leads NMS by $0.0$
to $0.4$ points; the $95\%$ interval includes zero on four scenes, and on $27$ of
the $40$ scene--seed pairs the two cover exactly the same errors. We read this as a practical tie, not a tested equivalence; the marginal-gain rule adds a guarantee rather than measured coverage. Dispersion
beats greedy coverage, peak uncertainty and stratified sampling on every seed of
every scene, with intervals that exclude zero; even its smallest lead, over
stratified sampling on HongHu, is $0.7$ points ($[0.5, 0.9]$). Stratified sampling, which forces
one site into each part of the labelled region, covers $0.7$ to $26.6$ points
fewer errors than dispersion: breadth imposed for its own sake costs coverage.

The scene-wide baselines cover little: on KSC the lawnmower's first 15 footprints
fall entirely on unlabelled ground, so it covers no errors, and on every scene but
Indian Pines both scene-wide baselines stay at or below $3\%$. On the sparsely
labelled scenes part of this gap reflects where the labels are rather than where
the classifier is uncertain, so the region-matched baselines are the fairer
reference. Restricted to the labelled region, the best uncertainty-agnostic
baseline covers $30.7\%$ of the errors on Indian Pines (the lawnmower, which suits
its wall-to-wall labelling), $21.3\%$ on KSC (random), and $0.7$ to $7.6\%$ on the
other scenes. Dispersion covers $1.7$ to $18$ times as many errors as the best of
them (Indian Pines $53.0$ against $30.7\%$, LongKou $27.2$ against $1.5\%$), or
$3.3$ to $38.1$ percentage points more, and it does so on every seed. The random
baselines are averages over $20$ draws per classifier seed; a single draw scatters
around that average with a standard deviation of up to $5$ points (random,
labelled, on KSC), so a comparison against one draw would be noisy. Knowing where
the labelled land cover is helps the baselines, but knowing where the classifier
is uncertain helps much more.

Table~\ref{tab:precision} gives a complementary view, \emph{inspection
precision}. The small error coverage on the finest-resolution scenes (HanChuan
$6.6\%$, HongHu $4.0\%$) is a budget-versus-scale effect rather than a failure of
guidance: there, precision is highest, $80\%$ for dispersion on HongHu, because
errors fragment into many small clusters across a large scene, so 15 footprints
cover little of the total error mass while landing mostly on true errors. Greedy
coverage has the highest precision on every scene, $0.2$ to $13.5$ points above
dispersion: its mean-uncertainty score favours footprints dense in ambiguous
pixels, which dispersion trades for more covered mass. It exceeds every baseline
on every scene. On Indian Pines, where labels are dense, the lawnmower
($43.9\%$) is close to dispersion ($45.4\%$) but below greedy coverage ($49.3\%$).
Stratified sampling has lower precision than dispersion on every scene.

\begin{table*}[t]
\caption{\textbf{Error coverage (\%):} the percentage of genuine classifier errors
  (majority-vote misclassifications against ground truth) that fall within an
  inspected footprint, with a budget of 15 sites. Mean $\pm$ std over seeds
  (Indian Pines: 10, others: 5).
  Scene-wide baselines sweep the whole scene; region-matched baselines are
  restricted to the labelled region (Section~\ref{sec:protocol}); random
  baselines are averaged over $20$ draws per classifier seed. The best policy per
  scene is bolded. Dispersion and NMS differ by at most $0.4$ points (paired
  comparison in Section~\ref{sec:error_reduction}).}
   \vspace{-0.3cm}
  \label{tab:error_reduction}
  \small
  \setlength{\tabcolsep}{1pt}
  \begin{tabular}{l cccc c ccc}
    \toprule
    & \multicolumn{4}{c}{Archival airborne} & & \multicolumn{3}{c}{UAV-borne (WHU-Hi)} \\
    \cmidrule(lr){2-5} \cmidrule(lr){7-9}
    Strategy & Indian Pines & Salinas & KSC & Pavia U. & & LongKou & HanChuan & HongHu \\
    \midrule
    Peak uncertainty      & $32.4 \pm 2.7$ & $5.6 \pm 1.1$ & $27.4 \pm 3.9$ & $7.6 \pm 1.4$ & & $6.7 \pm 1.4$ & $3.1 \pm 0.3$ & $2.5 \pm 0.3$ \\
    Greedy coverage       & $42.4 \pm 2.5$ & $32.5 \pm 1.9$ & $54.9 \pm 2.3$ & $23.3 \pm 0.8$ & & $18.8 \pm 1.4$ & $5.0 \pm 0.2$ & $3.1 \pm 0.2$ \\
    Unc.-wt.\ dispersion  & $\mathbf{53.0 \pm 1.9}$ & $\mathbf{45.7 \pm 1.3}$ & $\mathbf{56.7 \pm 2.3}$ & $\mathbf{31.8 \pm 1.0}$ & & $\mathbf{27.2 \pm 0.7}$ & $\mathbf{6.6 \pm 0.1}$ & $\mathbf{4.0 \pm 0.1}$ \\
    Unc.-ranked NMS       & $52.9 \pm 2.2$ & $45.3 \pm 1.2$ & $56.4 \pm 2.5$ & $31.6 \pm 0.7$ & & $26.9 \pm 0.7$ & $\mathbf{6.6 \pm 0.1}$ & $\mathbf{4.0 \pm 0.1}$ \\
    Stratified unc.       & $43.0 \pm 1.5$ & $19.0 \pm 1.2$ & $40.6 \pm 2.7$ & $13.0 \pm 1.0$ & & $14.6 \pm 1.1$ & $4.9 \pm 0.1$ & $3.3 \pm 0.1$ \\
    \midrule
    Lawnmower, scene-wide & $30.7 \pm 0.9$ & $1.7 \pm 0.2$ & $0.0 \pm 0.0$ & $0.5 \pm 0.0$ & & $0.5 \pm 0.0$ & $0.3 \pm 0.0$ & $0.2 \pm 0.0$ \\
    Random, scene-wide    & $14.8 \pm 1.1$ & $3.0 \pm 0.6$ & $1.0 \pm 0.2$ & $1.7 \pm 0.2$ & & $1.4 \pm 0.1$ & $1.0 \pm 0.1$ & $0.7 \pm 0.0$ \\
    \midrule
    Random, labelled      & $21.1 \pm 0.8$ & $5.4 \pm 0.6$ & $21.3 \pm 1.4$ & $4.8 \pm 0.4$ & & $1.5 \pm 0.2$ & $0.9 \pm 0.1$ & $0.7 \pm 0.0$ \\
    Lawnmower, labelled   & $30.7 \pm 0.9$ & $1.8 \pm 0.2$ & $9.9 \pm 0.5$ & $1.1 \pm 0.1$ & & $0.5 \pm 0.0$ & $0.3 \pm 0.0$ & $0.2 \pm 0.0$ \\
    Systematic, labelled  & $20.2 \pm 0.8$ & $7.6 \pm 0.2$ & $5.3 \pm 0.5$ & $2.3 \pm 0.2$ & & $1.5 \pm 0.1$ & $1.2 \pm 0.0$ & $0.6 \pm 0.0$ \\
    \bottomrule
  \end{tabular}
\end{table*}

\begin{table*}[t]
\caption{\textbf{Inspection precision (\%):} of the labelled pixels an inspection
  visits, the fraction that are genuine classifier errors. Mean $\pm$ std over
  seeds. On the finest-resolution scenes (HanChuan, HongHu) the error \emph{mass}
  a 15-site budget can cover is small (Table~\ref{tab:error_reduction}), yet
  precision is highest there because errors fragment across a large scene and the
  uncertainty-driven policies still land mostly on true errors. The best
  uncertainty-driven policy per scene is bolded. Random baselines are averaged
  over $20$ draws per classifier seed.}
   \vspace{-0.3cm}
  \label{tab:precision}
  \small
  \setlength{\tabcolsep}{2pt}
  \begin{tabular}{l cccc c ccc}
    \toprule
    & \multicolumn{4}{c}{Archival airborne} & & \multicolumn{3}{c}{UAV-borne (WHU-Hi)} \\
    \cmidrule(lr){2-5} \cmidrule(lr){7-9}
    Strategy & Indian Pines & Salinas & KSC & Pavia U. & & LongKou & HanChuan & HongHu \\
    \midrule
    Peak uncertainty      & $41.6 \pm 4.1$ & $8.8 \pm 2.4$ & $35.5 \pm 4.2$ & $18.3 \pm 3.3$ & & $12.2 \pm 2.1$ & $42.1 \pm 3.1$ & $71.6 \pm 7.2$ \\
    Greedy coverage       & $\mathbf{49.3 \pm 2.1}$ & $\mathbf{52.4 \pm 2.1}$ & $\mathbf{45.2 \pm 2.8}$ & $\mathbf{44.9 \pm 3.3}$ & & $\mathbf{53.3 \pm 3.2}$ & $\mathbf{69.8 \pm 2.1}$ & $\mathbf{80.3 \pm 2.6}$ \\
    Unc.-wt.\ dispersion  & $45.4 \pm 1.8$ & $44.9 \pm 1.5$ & $43.4 \pm 1.3$ & $38.7 \pm 1.7$ & & $39.8 \pm 1.7$ & $65.0 \pm 0.8$ & $80.1 \pm 1.4$ \\
    Unc.-ranked NMS       & $45.2 \pm 2.1$ & $44.2 \pm 1.3$ & $43.1 \pm 1.1$ & $38.3 \pm 1.3$ & & $39.2 \pm 1.8$ & $65.0 \pm 0.8$ & $80.1 \pm 1.4$ \\
    Stratified unc.       & $37.2 \pm 1.7$ & $21.8 \pm 1.0$ & $25.0 \pm 1.2$ & $22.3 \pm 2.1$ & & $23.2 \pm 2.3$ & $55.9 \pm 1.4$ & $69.8 \pm 3.3$ \\
    \midrule
    Lawnmower, scene-wide & $43.9 \pm 1.4$ & $4.3 \pm 0.5$ & $0.0 \pm 0.0$ & $5.4 \pm 0.5$ & & $0.8 \pm 0.1$ & $4.2 \pm 0.2$ & $7.1 \pm 0.3$ \\
    Random, scene-wide    & $22.3 \pm 1.6$ & $5.5 \pm 1.3$ & $21.7 \pm 5.0$ & $9.9 \pm 1.5$ & & $2.1 \pm 0.2$ & $13.6 \pm 1.7$ & $17.4 \pm 1.0$ \\
    \midrule
    Random, labelled      & $21.3 \pm 0.6$ & $5.9 \pm 0.7$ & $19.0 \pm 1.2$ & $8.9 \pm 0.6$ & & $2.1 \pm 0.2$ & $10.8 \pm 0.9$ & $15.6 \pm 1.0$ \\
    Lawnmower, labelled   & $43.9 \pm 1.4$ & $2.6 \pm 0.3$ & $24.3 \pm 1.4$ & $5.3 \pm 0.6$ & & $0.8 \pm 0.1$ & $3.6 \pm 0.2$ & $5.2 \pm 0.2$ \\
    Systematic, labelled  & $22.3 \pm 1.0$ & $8.8 \pm 0.3$ & $16.8 \pm 1.7$ & $9.5 \pm 0.8$ & & $2.2 \pm 0.1$ & $16.1 \pm 0.4$ & $13.1 \pm 0.4$ \\
    \bottomrule
  \end{tabular}
\end{table*}

\subsection{Spatially Blocked Evaluation}
\label{sec:blocked_eval}

The comparison above plans over the whole labelled region with surfaces whose
members each trained on a random tenth of it, so many inspected pixels border
training pixels. Under the blocked protocol of Section~\ref{sec:error_corr} we
instead restrict both planning and scoring to the labelled pixels of the test
blocks: the surface is renormalized over them, every policy selects only there,
the region-matched baselines are restricted the same way, and only errors there
count. Table~\ref{tab:blocked} reports error coverage.

The ordering of the main comparison holds. Dispersion and NMS again cover the most
held-out errors, and they differ by at most $1.3$ points in either direction.
Dispersion covers $1.4$ to $11.2$ times as many held-out errors as the best
region-matched baseline, $6.0$ to $30.3$ points more, wins on every seed of every
scene, and has paired $95\%$ intervals that exclude zero. Its margin is smallest
on Indian Pines ($68.0$ against $48.7\%$ for the labelled lawnmower), whose dense
labels suit a systematic sweep. Percentages are higher than in
Table~\ref{tab:error_reduction} because the test region holds only $36$ to $46\%$
of the labelled pixels, so 15 footprints reach a larger share of it, and seed
spreads are larger because a handful of blocks decides what the test region
contains. Inspection precision is less uniform under this protocol: on Indian
Pines the labelled lawnmower is the most precise ($86.7\%$) and on KSC peak
uncertainty ($86.0\%$), while dispersion, level with NMS, leads only on HongHu
($78.0\%$). The
advantage in error coverage, which is the quantity the planner targets, therefore
survives the removal of spatial leakage, although disagreement ranks errors less
well there (Table~\ref{tab:auc}).

\begin{table*}[t]
\caption{\textbf{Spatially blocked evaluation: error coverage of held-out errors
  (\%).} Planning and scoring are restricted to the labelled pixels of test blocks
  ($32\times32$-pixel blocks, none within $7$ pixels of a training block); the
  surface is renormalized over them. Mean $\pm$ std over seeds (Indian Pines: 10,
  others: 5); random baselines averaged over $20$ draws per seed. The best policy
  per scene is bolded.}
   \vspace{-0.3cm}
  \label{tab:blocked}
  \small
  \setlength{\tabcolsep}{1pt}
  \begin{tabular}{l cccc c ccc}
    \toprule
    & \multicolumn{4}{c}{Archival airborne} & & \multicolumn{3}{c}{UAV-borne (WHU-Hi)} \\
    \cmidrule(lr){2-5} \cmidrule(lr){7-9}
    Strategy & Indian Pines & Salinas & KSC & Pavia U. & & LongKou & HanChuan & HongHu \\
    \midrule
    Peak uncertainty      & $50.2 \pm 5.4$ & $12.3 \pm 3.2$ & $39.8 \pm 10.4$ & $12.6 \pm 4.5$ & & $13.2 \pm 4.9$ & $4.0 \pm 1.2$ & $4.6 \pm 0.7$ \\
    Greedy coverage       & $53.9 \pm 7.4$ & $23.6 \pm 5.0$ & $68.1 \pm 9.7$ & $26.9 \pm 7.3$ & & $25.6 \pm 9.2$ & $8.8 \pm 0.6$ & $6.2 \pm 0.5$ \\
    Unc.-wt.\ dispersion  & $\mathbf{68.0 \pm 5.7}$ & $33.0 \pm 4.8$ & $\mathbf{75.6 \pm 11.5}$ & $34.0 \pm 10.9$ & & $33.3 \pm 5.3$ & $\mathbf{11.6 \pm 1.0}$ & $\mathbf{7.8 \pm 0.2}$ \\
    Unc.-ranked NMS       & $67.7 \pm 4.6$ & $\mathbf{34.1 \pm 5.3}$ & $74.3 \pm 10.1$ & $\mathbf{34.1 \pm 10.8}$ & & $\mathbf{33.4 \pm 5.4}$ & $\mathbf{11.6 \pm 1.1}$ & $\mathbf{7.8 \pm 0.2}$ \\
    Stratified unc.       & $59.7 \pm 5.1$ & $17.5 \pm 6.7$ & $66.3 \pm 4.9$ & $19.9 \pm 3.1$ & & $20.0 \pm 5.0$ & $7.8 \pm 1.0$ & $6.2 \pm 0.3$ \\
    \midrule
    Random, labelled      & $48.4 \pm 3.1$ & $12.7 \pm 1.2$ & $48.1 \pm 3.3$ & $11.0 \pm 2.0$ & & $2.7 \pm 0.4$ & $2.2 \pm 0.2$ & $1.7 \pm 0.1$ \\
    Lawnmower, labelled   & $48.7 \pm 8.4$ & $4.9 \pm 4.2$ & $20.8 \pm 8.6$ & $0.7 \pm 0.7$ & & $2.9 \pm 1.4$ & $1.1 \pm 1.1$ & $0.5 \pm 0.2$ \\
    Systematic, labelled  & $34.3 \pm 3.0$ & $9.6 \pm 2.4$ & $20.4 \pm 3.7$ & $6.3 \pm 3.7$ & & $3.0 \pm 2.5$ & $1.8 \pm 0.5$ & $1.1 \pm 0.6$ \\
    \bottomrule
  \end{tabular}
\end{table*}

\paragraph{Block size.}
A $32$-pixel block is $640$\,m wide on Indian Pines but $1.4$\,m on HongHu, so
whether it is large enough to break spatial correlation depends on the scene. We
therefore repeated the blocked protocol with $64$- and $128$-pixel blocks, keeping
the $7$-pixel buffer (Table~\ref{tab:blocksize}). On Indian Pines with $128$-pixel
blocks only $3\%$ of the labelled pixels qualify as test pixels, because a block is
nearly as large as the $145$-pixel scene, and we omit that setting. Larger blocks
lower the AUC on most scenes: on LongKou from $0.792$ to $0.705$, and on the
archival scenes to $0.56$ to $0.62$ at $128$ pixels. On HanChuan and HongHu, where
even a $128$-pixel block is only $14$ and $5.5$\,m wide, it stays near $0.72$ and
$0.75$. Dispersion covers more held-out errors than the best region-matched
baseline at every block size, $1.3$ to $11.2$ times as many, and NMS stays within
$1.3$ points of it. The ratio does shrink as blocks grow on several scenes
(LongKou from $11.2$ to $2.9$, Salinas from $2.6$ to $1.3$), so the farther the
test ground lies from any training pixel, the weaker both the error ranking and
the planning advantage become, although neither disappears.

\begin{table}[t]
  \caption{Block-size sweep of the blocked protocol: block side in pixels and
  metres, test share of the labelled pixels, AUC of $u$ on test pixels, error
  coverage (\%) of dispersion and of the best region-matched (RM) baseline inside
  the test blocks, and their ratio. Mean over seeds (Indian Pines: 10, others: 5).
  Indian Pines at $128$ pixels (test share $3\%$) is omitted.}
   \vspace{-0.3cm}
  \label{tab:blocksize}
  \small
  \setlength{\tabcolsep}{3pt}
  \begin{tabular}{l rr r c cc r}
    \toprule
    Scene & px & m & Test & AUC & Disp. & Best RM & Ratio \\
    \midrule
    Indian Pines & $32$ & $640$ & $36\%$ & $0.630$ & $68.0$ & $48.7$ & $1.4$ \\
                 & $64$ & $1280$ & $41\%$ & $0.574$ & $63.7$ & $48.5$ & $1.3$ \\
    \midrule
    Salinas & $32$ & $118$ & $40\%$ & $0.648$ & $33.0$ & $12.7$ & $2.6$ \\
            & $64$ & $237$ & $49\%$ & $0.703$ & $21.8$ & $10.0$ & $2.2$ \\
            & $128$ & $474$ & $42\%$ & $0.590$ & $15.5$ & $12.3$ & $1.3$ \\
    \midrule
    KSC & $32$ & $576$ & $46\%$ & $0.692$ & $75.6$ & $48.1$ & $1.6$ \\
        & $64$ & $1152$ & $50\%$ & $0.671$ & $68.9$ & $46.6$ & $1.5$ \\
        & $128$ & $2304$ & $51\%$ & $0.624$ & $67.6$ & $43.6$ & $1.6$ \\
    \midrule
    Pavia U. & $32$ & $42$ & $40\%$ & $0.652$ & $34.0$ & $11.0$ & $3.1$ \\
             & $64$ & $83$ & $49\%$ & $0.655$ & $26.7$ & $8.9$ & $3.0$ \\
             & $128$ & $166$ & $50\%$ & $0.558$ & $21.7$ & $11.1$ & $1.9$ \\
    \midrule
    LongKou & $32$ & $15$ & $39\%$ & $0.792$ & $33.3$ & $3.0$ & $11.2$ \\
            & $64$ & $30$ & $50\%$ & $0.752$ & $22.1$ & $2.8$ & $8.0$ \\
            & $128$ & $59$ & $53\%$ & $0.705$ & $13.5$ & $4.7$ & $2.9$ \\
    \midrule
    HanChuan & $32$ & $3.5$ & $40\%$ & $0.740$ & $11.6$ & $2.2$ & $5.3$ \\
             & $64$ & $7.0$ & $50\%$ & $0.719$ & $7.6$ & $1.8$ & $4.2$ \\
             & $128$ & $14$ & $53\%$ & $0.717$ & $5.8$ & $1.9$ & $3.0$ \\
    \midrule
    HongHu & $32$ & $1.4$ & $40\%$ & $0.748$ & $7.8$ & $1.7$ & $4.5$ \\
           & $64$ & $2.8$ & $49\%$ & $0.750$ & $5.2$ & $1.9$ & $2.7$ \\
           & $128$ & $5.5$ & $53\%$ & $0.746$ & $5.4$ & $1.5$ & $3.6$ \\
    \bottomrule
  \end{tabular}
\end{table}

\subsection{Uncertainty Coverage and Spatial Structure}
\label{sec:why}

Dispersion and NMS cover more errors because they cover more of the scene's
\emph{uncertainty mass}, which Section~\ref{sec:error_corr} shows coincides with
genuine error. Table~\ref{tab:reduction} reports the share of each scene's
uncertainty mass that the 15 footprints cover, the quantity
objective~\eqref{eq:objective} maximizes.

\begin{table*}[!tbp]
 \caption{Uncertainty covered (\%): the share of the scene's uncertainty mass
  inside the inspected footprints, objective~\eqref{eq:objective}, with 15 sites.
  Mean $\pm$ std over seeds (Indian Pines: 10; all others: 5). Every
  uncertainty-driven policy exceeds every baseline on every scene, and dispersion
  and NMS are the best of them. The scene-wide baselines cover little except on
  the densely labelled Indian Pines.}
   \vspace{-0.3cm}
  \label{tab:reduction}
  \small
  \setlength{\tabcolsep}{1pt}
  \begin{tabular}{l cccc c ccc}
    \toprule
    & \multicolumn{4}{c}{Archival airborne} & & \multicolumn{3}{c}{UAV-borne (WHU-Hi)} \\
    \cmidrule(lr){2-5} \cmidrule(lr){7-9}
    Strategy & Indian Pines & Salinas & KSC & Pavia U. & & LongKou & HanChuan & HongHu \\
    \midrule
    Peak uncertainty      & $28.9 \pm 1.8$ & $8.0 \pm 1.0$ & $26.0 \pm 2.5$ & $7.5 \pm 1.3$ & & $6.3 \pm 1.3$ & $2.9 \pm 0.3$ & $2.2 \pm 0.2$ \\
    Greedy coverage       & $36.2 \pm 2.0$ & $20.6 \pm 0.8$ & $48.1 \pm 1.8$ & $18.1 \pm 1.2$ & & $10.7 \pm 0.7$ & $4.9 \pm 0.3$ & $3.2 \pm 0.2$ \\
    Unc.-wt.\ dispersion  & $\mathbf{45.4 \pm 0.9}$ & $\mathbf{30.5 \pm 0.7}$ & $\mathbf{50.1 \pm 1.5}$ & $\mathbf{26.0 \pm 0.3}$ & & $\mathbf{17.5 \pm 0.2}$ & $\mathbf{6.6 \pm 0.1}$ & $\mathbf{4.0 \pm 0.0}$ \\
    Unc.-ranked NMS       & $45.3 \pm 1.0$ & $30.3 \pm 0.7$ & $\mathbf{50.1 \pm 1.5}$ & $\mathbf{26.0 \pm 0.3}$ & & $17.4 \pm 0.2$ & $\mathbf{6.6 \pm 0.1}$ & $\mathbf{4.0 \pm 0.0}$ \\
    Stratified unc.       & $37.5 \pm 1.2$ & $16.8 \pm 0.8$ & $40.9 \pm 1.6$ & $12.7 \pm 0.3$ & & $11.1 \pm 0.3$ & $4.9 \pm 0.2$ & $3.2 \pm 0.1$ \\
    \midrule
    Lawnmower, scene-wide & $25.6 \pm 0.7$ & $1.6 \pm 0.1$ & $0.0 \pm 0.0$ & $0.8 \pm 0.1$ & & $0.5 \pm 0.0$ & $0.3 \pm 0.0$ & $0.3 \pm 0.0$ \\
    Random, scene-wide    & $15.1 \pm 0.7$ & $3.0 \pm 0.5$ & $1.1 \pm 0.2$ & $1.7 \pm 0.2$ & & $1.4 \pm 0.1$ & $0.9 \pm 0.1$ & $0.7 \pm 0.0$ \\
    \midrule
    Random, labelled      & $21.5 \pm 0.5$ & $5.2 \pm 0.5$ & $22.0 \pm 0.9$ & $4.6 \pm 0.3$ & & $1.4 \pm 0.1$ & $0.9 \pm 0.1$ & $0.7 \pm 0.0$ \\
    Lawnmower, labelled   & $25.6 \pm 0.7$ & $1.9 \pm 0.2$ & $9.4 \pm 0.8$ & $1.4 \pm 0.1$ & & $0.5 \pm 0.0$ & $0.3 \pm 0.0$ & $0.4 \pm 0.0$ \\
    Systematic, labelled  & $22.8 \pm 0.5$ & $5.9 \pm 0.3$ & $6.9 \pm 0.2$ & $2.6 \pm 0.1$ & & $1.5 \pm 0.1$ & $1.2 \pm 0.0$ & $0.8 \pm 0.0$ \\
    \bottomrule
  \end{tabular}
\end{table*}

Dispersion covers the most uncertainty on every scene, level with NMS (paired
difference $0.0$ to $0.1$ points), and it beats greedy coverage, peak uncertainty
and stratified sampling on every seed; its smallest paired leads are $0.8$ points
over greedy coverage and over stratified sampling on HongHu, with intervals that
exclude zero. Its lead over the best region-matched baseline is $3.2$ to $28.1$
points. Every uncertainty-driven policy covers more uncertainty than every
baseline. The standard deviations are small relative to the gaps between
policies. The magnitude is strongly scene-dependent: dispersion covers $50.1\%$
of the uncertainty mass on KSC and $45.4\%$ on Indian Pines but $4$ to $7\%$ on
the finest-resolution UAV scenes (HanChuan and HongHu), where ambiguity fragments
into many small boundaries spread over a large scene that 15 footprints can only
partly reach. The ranking is unchanged; only the attainable magnitude scales with
the spatial granularity of uncertainty.

\begin{figure*}[t]
  \centering
  \includegraphics[width=0.76\linewidth]{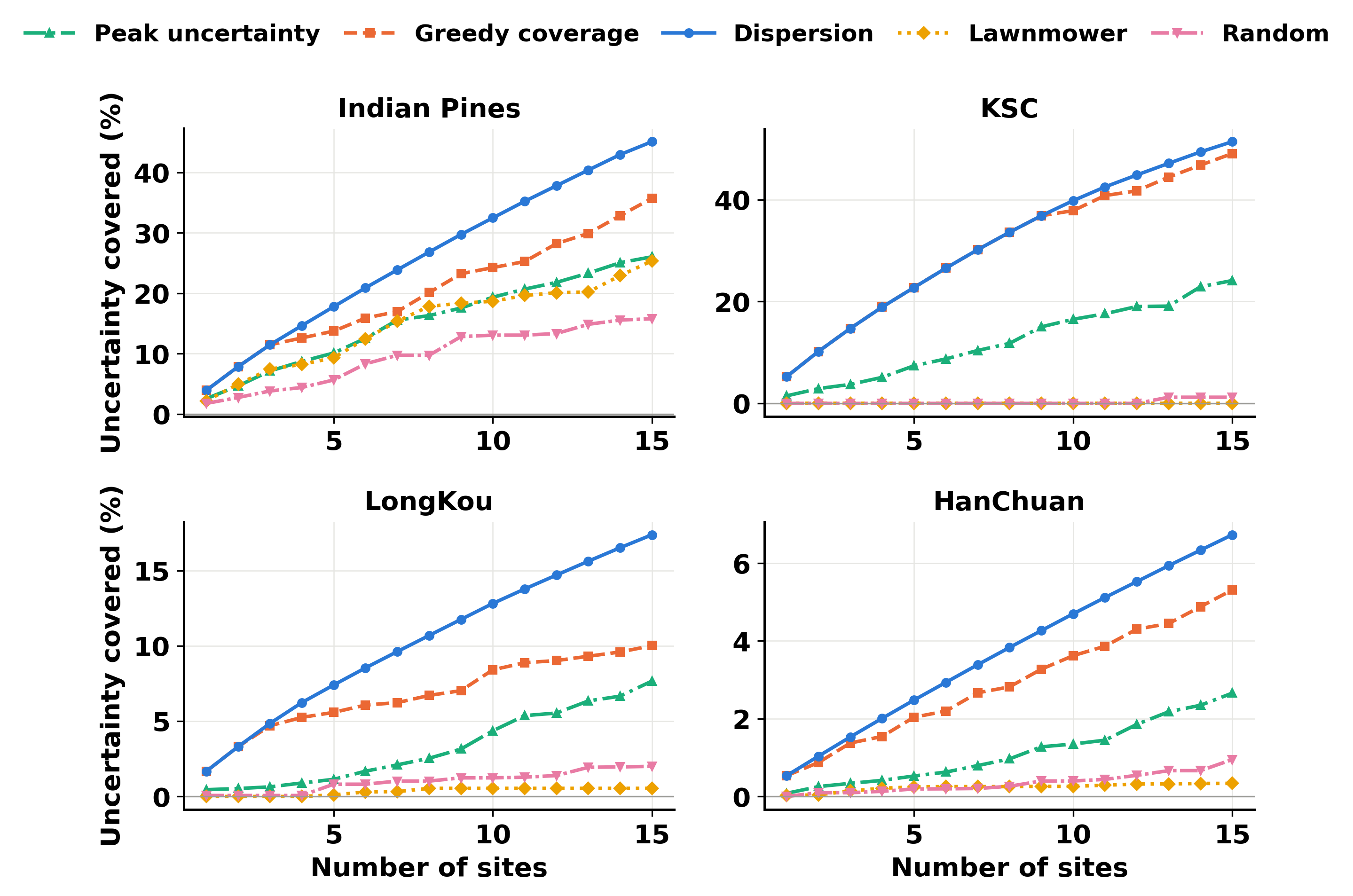}
   \vspace{-0.4cm}
  \caption{Uncertainty covered as inspection sites are added, for four
  representative scenes (seed 0). The dispersion policy leads or ties at every budget on each scene shown and is still rising at the 15th site; the scene-wide
  lawnmower and random baselines stay far below it. Note the differing vertical
  scales: attainable coverage is far larger on the coarser-resolution scenes
  (Indian Pines, KSC) than on the fine-grained UAV scenes.}
  \label{fig:reduction_curves}
  \Description{Four line plots of percent uncertainty covered against the number of sites, 1 to 15, for five policies on Indian Pines, KSC, LongKou and HanChuan; dispersion is highest or tied at every budget.}
\end{figure*}

Figure~\ref{fig:reduction_curves} traces how covered uncertainty accrues as the
budget is spent, for four representative scenes. Dispersion leads or ties at every budget on every one of them, and its curve is still rising steeply at the 15th site: the
budget, not the surface, limits coverage, as the budget sweep of
Section~\ref{sec:tau} confirms.

Table~\ref{tab:coverage} reports connected-anomaly coverage, the fraction of
spatially contiguous high-uncertainty regions that are inspected. Here the
picture is scene-dependent.

\begin{table*}[t]
  \caption{Connected-anomaly coverage (\%), mean $\pm$ std over seeds (Indian
  Pines: 10; all others: 5). The best uncertainty-driven policy per scene is
  bolded; unlike Table~\ref{tab:reduction}, it is not the same policy on every
  scene. Dispersion or NMS leads on five scenes and peak uncertainty on Salinas
  and Pavia University. Components are the 4-connected regions of $U \ge 0.7$.}
   \vspace{-0.3cm}
  \label{tab:coverage}
  \small
  \setlength{\tabcolsep}{1pt}
  \begin{tabular}{l cccc c ccc}
    \toprule
    & \multicolumn{4}{c}{Archival airborne} & & \multicolumn{3}{c}{UAV-borne (WHU-Hi)} \\
    \cmidrule(lr){2-5} \cmidrule(lr){7-9}
    Strategy & Indian Pines & Salinas & KSC & Pavia U. & & LongKou & HanChuan & HongHu \\
    \midrule
    Peak uncertainty      & $26.0 \pm 1.3$ & $\mathbf{24.1 \pm 4.0}$ & $27.1 \pm 1.2$ & $\mathbf{20.8 \pm 2.4}$ & & $10.9 \pm 1.3$ & $3.2 \pm 0.3$ & $2.0 \pm 0.1$ \\
    Greedy coverage       & $30.0 \pm 2.2$ & $6.5 \pm 1.7$ & $38.9 \pm 0.5$ & $4.1 \pm 1.9$ & & $4.2 \pm 0.8$ & $4.0 \pm 0.1$ & $2.3 \pm 0.3$ \\
    Unc.-wt.\ dispersion  & $\mathbf{40.0 \pm 1.5}$ & $14.0 \pm 2.8$ & $\mathbf{41.5 \pm 1.3}$ & $7.3 \pm 1.3$ & & $12.6 \pm 0.6$ & $\mathbf{5.7 \pm 0.2}$ & $\mathbf{2.9 \pm 0.2}$ \\
    Unc.-ranked NMS       & $39.7 \pm 1.5$ & $14.2 \pm 2.9$ & $41.4 \pm 1.3$ & $7.3 \pm 1.3$ & & $\mathbf{13.0 \pm 1.0}$ & $\mathbf{5.7 \pm 0.2}$ & $\mathbf{2.9 \pm 0.2}$ \\
    Stratified unc.       & $34.8 \pm 3.0$ & $22.2 \pm 3.1$ & $36.6 \pm 1.3$ & $14.5 \pm 2.4$ & & $11.9 \pm 1.5$ & $4.8 \pm 0.5$ & $2.5 \pm 0.2$ \\
    \midrule
    Lawnmower, scene-wide & $22.9 \pm 1.6$ & $4.0 \pm 1.2$ & $0.0 \pm 0.0$ & $3.6 \pm 0.5$ & & $1.0 \pm 0.3$ & $0.2 \pm 0.0$ & $0.3 \pm 0.0$ \\
    Random, scene-wide    & $15.3 \pm 0.6$ & $3.0 \pm 0.3$ & $1.1 \pm 0.3$ & $1.7 \pm 0.2$ & & $1.4 \pm 0.1$ & $0.9 \pm 0.1$ & $0.7 \pm 0.0$ \\
    \midrule
    Random, labelled      & $21.3 \pm 0.5$ & $3.9 \pm 0.5$ & $21.0 \pm 0.7$ & $3.3 \pm 0.3$ & & $1.3 \pm 0.1$ & $0.9 \pm 0.0$ & $0.7 \pm 0.0$ \\
    Lawnmower, labelled   & $22.9 \pm 1.6$ & $4.5 \pm 1.4$ & $10.3 \pm 1.0$ & $6.4 \pm 1.0$ & & $1.0 \pm 0.3$ & $0.2 \pm 0.1$ & $0.3 \pm 0.0$ \\
    Systematic, labelled  & $23.5 \pm 1.3$ & $4.1 \pm 0.8$ & $6.9 \pm 0.7$ & $7.5 \pm 0.7$ & & $2.3 \pm 0.2$ & $1.3 \pm 0.1$ & $0.8 \pm 0.1$ \\
    \bottomrule
  \end{tabular}
\end{table*}

Dispersion, level with NMS, covers the most connected anomalies on Indian Pines
($40.0\%$), KSC ($41.5\%$), HanChuan ($5.7\%$) and HongHu ($2.9\%$), and on
LongKou NMS ($13.0\%$) and dispersion ($12.6\%$) lead peak uncertainty
($10.9\%$). Peak uncertainty wins on Salinas ($24.1$ against $14.0\%$ for
dispersion) and Pavia University ($20.8$ against $7.3\%$). Stratified sampling is
second on those two scenes ($22.2$ and $14.5\%$) but still trails peak. Greedy
coverage trails dispersion on every scene and falls far behind on Salinas, Pavia
University and LongKou. The same winners result when components are weighted by
their size and when we count the share of components an inspection touches at
all. The best policy for \emph{covering uncertainty} and \emph{covering errors}
is the same on all seven scenes, but the best policy for \emph{covering connected
anomalies} depends on scene structure and, as Section~\ref{sec:tau} shows, on the
threshold that defines an anomaly.

\begin{figure*}[t]
  \centering
  \includegraphics[width=0.84\linewidth]{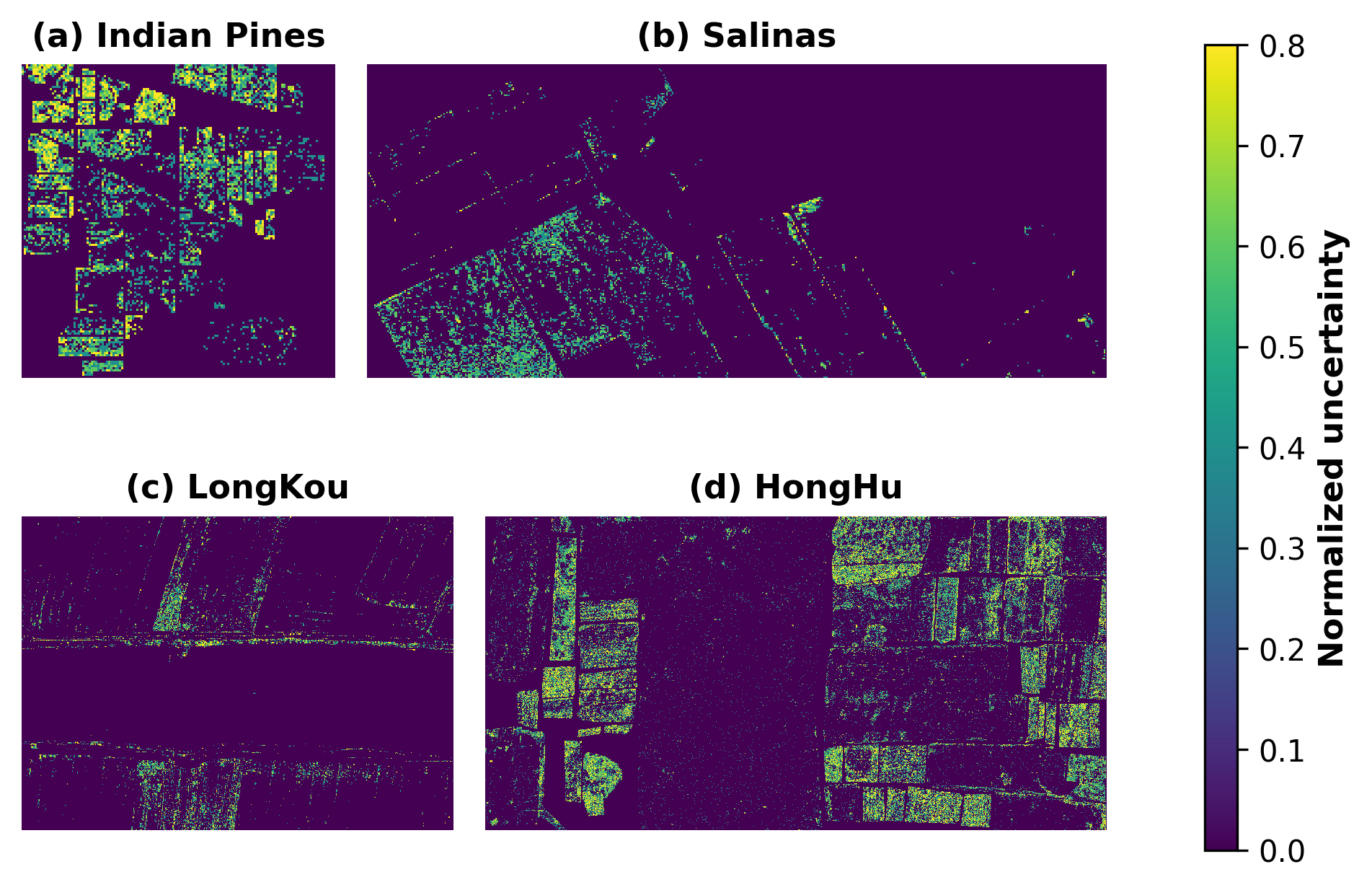}
   \vspace{-0.7cm}
\caption{Representative ensemble-disagreement uncertainty surfaces (seed 0,
  shared colour scale clipped at 0.8; unlabelled background masked to zero;
  (a) Indian Pines, (b) Salinas, (c) LongKou, (d) HongHu; Salinas, LongKou and
  HongHu are rotated by $90^\circ$ to fit the page). The arrangement
  differs between scenes: in Indian Pines and HongHu high uncertainty is speckled
  across many fields; in Salinas most of it lies in one or two fields, and in
  LongKou in thin strips along field edges. These arrangements shape
  connected-anomaly coverage (Table~\ref{tab:coverage}).}
  \label{fig:umaps}
  \Description{Four heatmaps of normalized classifier uncertainty on a shared colour scale for Indian Pines, Salinas, LongKou and HongHu, showing speckled, field-concentrated and edge-strip patterns.}
\end{figure*}

\begin{figure}[!ht]
  \centering
  \includegraphics[width=0.8\linewidth]{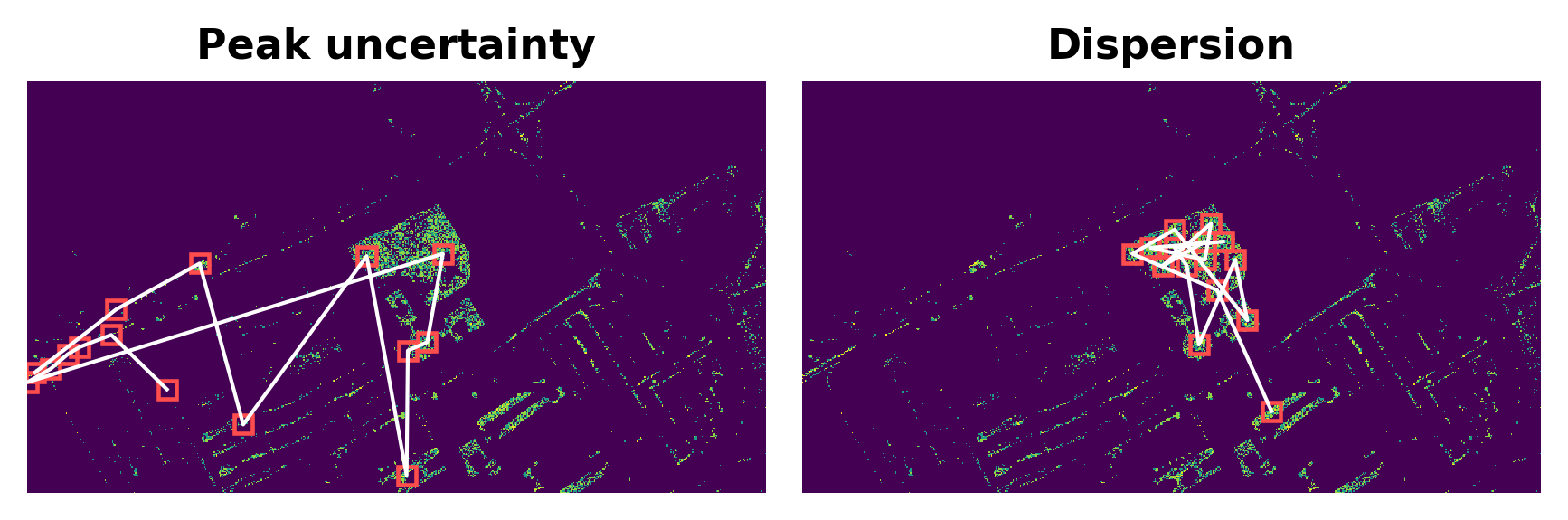}
    \vspace{-0.3cm}
  \caption{Two policies on the same Pavia University uncertainty surface (seed
  0). Dispersion (right) places most of its footprints on the one dense
  high-uncertainty patch, which holds most of the mass, while peak-uncertainty
  selection (left) centres each footprint on an isolated maximum and spreads
  across the scene. On Pavia University the spread-out policy covers more
  connected anomalies (Table~\ref{tab:coverage}), while dispersion covers more
  uncertainty mass and more errors (Tables~\ref{tab:reduction}
  and~\ref{tab:error_reduction}). Squares mark the sensor field of view; the
  scene is rotated by $90^\circ$ to fit the page, and lines join the sites in the order each policy selected them.}
  \Description{Two side-by-side heatmap panels of the same uncertainty surface
  with overlaid inspection paths: the dispersion policy clusters its waypoints on
  a dense patch, the peak policy spreads them across the scene.}
  \label{fig:policy_compare}
\end{figure}

The pattern is a trade-off between depth and breadth, but which policy plays
which role depends on the scene (Figures~\ref{fig:umaps}
and~\ref{fig:policy_compare}). Connected-anomaly coverage averages over hundreds
of small components, so it rewards reaching many separate components rather than
covering much mass. Dispersion goes where the mass is. When high uncertainty is
speckled across many fields, as in Indian Pines and the WHU-Hi scenes, following
the mass already spreads its sites, and it covers the most components too. When
most of the mass lies in one or two dense fields, as in Salinas and Pavia
University, dispersion stays there, while peak uncertainty centres each footprint
on an isolated maximum elsewhere and so reaches more separate components. The
waypoint spread of Table~\ref{tab:dispersion} follows the same split: peak spreads
its sites $1.5$ and $3.5$ times as widely as dispersion on Salinas and Pavia
University, the two scenes it wins, and no more widely than $1.1$ times on the
others. Breadth alone does not explain the win: stratified sampling spreads its
sites about as widely as random sampling (Table~\ref{tab:dispersion}) yet trails
peak uncertainty on both scenes, because peak centres each footprint on a local
maximum while stratified sampling takes the best footprint of a fixed stratum.
Section~\ref{sec:morphology} tests whether the split can be predicted
from the surface. In every case all uncertainty-driven policies still exceed the
baselines on uncertainty covered; the scene dependence is a choice
\emph{among} uncertainty-driven policies.

\subsection{Is the Best Strategy Predictable from Uncertainty Geometry?}
\label{sec:morphology}

The scene dependence of the connected-anomaly winner (Table~\ref{tab:coverage})
raises a spatial-analytics question: can we predict, from the \emph{geometry} of
the uncertainty surface alone, whether peak uncertainty or dispersion will cover
connected anomalies better? A
positive answer would turn the observation into an operating rule.

For each scene we thresholded the uncertainty surface, extracted connected
high-uncertainty components, and computed several descriptors of their spatial
organization: a Herfindahl-style concentration index
$C = \sum_i A_i^2 / (\sum_i A_i)^2$ over component areas $A_i$; the fraction of
high-uncertainty mass in components larger than one field-of-view footprint; and
component counts, each at several thresholds. We correlated each descriptor with
the margin between the two (dispersion minus peak connected-anomaly coverage) across the seven scenes.

With seven scenes this analysis is exploratory: it can show that an obvious
descriptor fails, not establish which one works. We compare each descriptor with
the margin measured on components defined at the \emph{same} threshold, since a
descriptor of the $0.6$ components says little about which policy covers the
$0.7$ components. Table~\ref{tab:morph} lists the margins and two descriptors.
The result is negative: \emph{no single scalar descriptor consistently predicts
the winner}. At $0.6$ the component count and the footprint-relative measure
correlate strongly and negatively with the margin (Pearson $r = -0.88$, Spearman
$\rho = -0.76$ and $-0.79$), but two scenes produce this: HanChuan and HongHu have
thousands of components at $0.6$, the only components larger than a footprint
($5$ to $6\%$ of the mass), and margins of $-0.4$ and $-0.3$ points, a near tie;
on the other five scenes dispersion wins by $7$ to $18$ points. The concentration
index correlates only moderately at $0.6$ ($r = +0.47$). At $0.7$ and $0.8$,
where the winner actually changes between scenes (peak wins on Salinas and Pavia
University by $10$ to $18$ points), no descriptor tracks the margin: the
concentration index gives $r = +0.07$ and $-0.56$, the component count $+0.03$
and $+0.12$, and the footprint-relative measure is zero on every scene. Almost no
connected high-uncertainty component is larger than one footprint, because in all
seven scenes high-uncertainty pixels lie in small fragments, speckle within
ambiguous fields or thin strips along field edges, whether the underlying
land-cover regions are large (Salinas, Pavia University) or small (Indian Pines).
The two scenes where peak wins at $0.7$ have the fewest components ($192$ and
$259$), but KSC, with the next fewest, is among dispersion's largest wins; on
concentration, Salinas is highest and Pavia University lies mid-range, between
LongKou and KSC. The
qualitative distinction the surfaces suggest, concentrated versus scattered
ambiguity, does not reduce at the level of thresholded components to a
one-dimensional index that separates the regimes. Which policy best covers
connected anomalies is therefore scene-dependent in a way a scalar measure of
concentration does not capture, and a practitioner should not expect a simple
geometric rule to choose the policy in advance. For uncertainty covered and error
coverage the choice is not scene-dependent: dispersion, or equivalently NMS, is
best on every scene (Tables~\ref{tab:reduction} and~\ref{tab:error_reduction}).

\begin{table}[t]
  \caption{Connected-anomaly margin (dispersion minus peak uncertainty, percentage
  points; mean over seeds) with components defined at three thresholds, and two
  descriptors of the $0.7$ components: their number $n$ and the concentration
  index $C$. A negative margin means peak uncertainty covers more components.}
   \vspace{-0.3cm}
  \label{tab:morph}
  \small
  \setlength{\tabcolsep}{4pt}
  \begin{tabular}{l ccc cc}
    \toprule
    & \multicolumn{3}{c}{Margin at threshold} & \multicolumn{2}{c}{Components at $0.7$} \\
    \cmidrule(lr){2-4} \cmidrule(lr){5-6}
    Scene & $0.6$ & $0.7$ & $0.8$ & $n$ & $C$ \\
    \midrule
    Indian Pines & $+11.6$ & $+14.1$ & $+15.4$ & $550$ & $0.0088$ \\
    Salinas      & $+17.9$ & $-10.1$ & $-15.2$ & $192$ & $0.0099$ \\
    KSC          & $+15.8$ & $+14.4$ & $+18.3$ & $355$ & $0.0071$ \\
    Pavia U.     & $+12.7$ & $-13.5$ & $-17.9$ & $259$ & $0.0049$ \\
    LongKou      & $+6.8$  & $+1.6$  & $-0.4$  & $616$ & $0.0024$ \\
    HanChuan     & $-0.4$  & $+2.5$  & $+4.5$  & $4{,}690$ & $0.0007$ \\
    HongHu       & $-0.3$  & $+1.0$  & $+2.1$  & $9{,}840$ & $0.0003$ \\
    \bottomrule
  \end{tabular}
\end{table}

\subsection{High-Uncertainty Coverage and Waypoint Dispersion}
\label{sec:coverage_dispersion}

Tables~\ref{tab:coverage_pix} and~\ref{tab:dispersion} report the two remaining
spatial metrics. \emph{High-uncertainty coverage}, the fraction of
labelled pixels with $U>0$ inside an inspected footprint, agrees with
uncertainty covered: dispersion and NMS are the best policies on every scene,
and every uncertainty-driven policy clears the scene-wide baselines,
which cover almost none of the high-uncertainty pixels (none on KSC, where the
lawnmower's footprints fall on unlabelled ground).

\begin{table*}[t]
  \caption{High-uncertainty coverage (\%), the fraction of labelled pixels with
  $U>0$ (the above-median pixels; the median is zero) falling within an inspected
  footprint, mean $\pm$ std over seeds (Indian Pines: 10; all others: 5). Higher
  is better. Dispersion and NMS are best on every scene.}
   \vspace{-0.3cm}
  \label{tab:coverage_pix}
  \small
  \setlength{\tabcolsep}{1pt}
  \begin{tabular}{l cccc c ccc}
    \toprule
    & \multicolumn{4}{c}{Archival airborne} & & \multicolumn{3}{c}{UAV-borne (WHU-Hi)} \\
    \cmidrule(lr){2-5} \cmidrule(lr){7-9}
    Strategy & Indian Pines & Salinas & KSC & Pavia U. & & LongKou & HanChuan & HongHu \\
    \midrule
    Peak uncertainty      & $26.7 \pm 1.6$ & $7.2 \pm 0.9$ & $25.0 \pm 2.5$ & $7.1 \pm 1.2$ & & $5.8 \pm 1.2$ & $2.7 \pm 0.3$ & $2.0 \pm 0.2$ \\
    Greedy coverage       & $33.8 \pm 1.9$ & $20.4 \pm 0.7$ & $47.1 \pm 1.5$ & $17.7 \pm 1.1$ & & $10.7 \pm 0.7$ & $4.3 \pm 0.2$ & $2.8 \pm 0.2$ \\
    Unc.-wt.\ dispersion  & $\mathbf{43.1 \pm 0.9}$ & $\mathbf{30.0 \pm 0.8}$ & $\mathbf{49.3 \pm 1.3}$ & $\mathbf{25.7 \pm 0.3}$ & & $\mathbf{17.2 \pm 0.3}$ & $\mathbf{6.0 \pm 0.0}$ & $\mathbf{3.5 \pm 0.0}$ \\
    Unc.-ranked NMS       & $43.0 \pm 0.9$ & $29.9 \pm 0.8$ & $\mathbf{49.3 \pm 1.3}$ & $25.6 \pm 0.3$ & & $17.1 \pm 0.3$ & $\mathbf{6.0 \pm 0.0}$ & $\mathbf{3.5 \pm 0.0}$ \\
    Stratified unc.       & $36.5 \pm 1.1$ & $16.4 \pm 0.8$ & $40.7 \pm 1.6$ & $12.4 \pm 0.3$ & & $10.8 \pm 0.3$ & $4.6 \pm 0.2$ & $2.9 \pm 0.1$ \\
    \midrule
    Lawnmower, scene-wide & $24.2 \pm 0.6$ & $1.5 \pm 0.1$ & $0.0 \pm 0.0$ & $0.7 \pm 0.1$ & & $0.5 \pm 0.0$ & $0.3 \pm 0.0$ & $0.3 \pm 0.0$ \\
    Random, scene-wide    & $15.1 \pm 0.6$ & $3.0 \pm 0.5$ & $1.1 \pm 0.2$ & $1.7 \pm 0.2$ & & $1.4 \pm 0.1$ & $0.9 \pm 0.1$ & $0.7 \pm 0.0$ \\
    \midrule
    Random, labelled      & $21.7 \pm 0.5$ & $5.2 \pm 0.5$ & $22.2 \pm 1.0$ & $4.6 \pm 0.3$ & & $1.4 \pm 0.1$ & $0.9 \pm 0.1$ & $0.7 \pm 0.0$ \\
    Lawnmower, labelled   & $24.2 \pm 0.6$ & $1.8 \pm 0.2$ & $9.7 \pm 0.8$ & $1.3 \pm 0.1$ & & $0.5 \pm 0.0$ & $0.4 \pm 0.0$ & $0.4 \pm 0.0$ \\
    Systematic, labelled  & $23.1 \pm 0.5$ & $6.0 \pm 0.2$ & $7.2 \pm 0.2$ & $2.5 \pm 0.1$ & & $1.4 \pm 0.1$ & $1.2 \pm 0.0$ & $0.8 \pm 0.0$ \\
    \bottomrule
  \end{tabular}
\end{table*}

\begin{table*}[t]
  \caption{Waypoint dispersion (mean pairwise distance between waypoints, in
  pixels), mean $\pm$ std over seeds. This is a \emph{descriptive} measure of how
  each policy distributes its waypoints, not a quality score: the most dispersed
  policy on every scene is a sampling design (random, systematic or stratified),
  and none of them covers the most errors.
  The lawnmower sweep is fixed by its grid.}
   \vspace{-0.3cm}
  \label{tab:dispersion}
  \small
  \setlength{\tabcolsep}{1pt}
  \begin{tabular}{l cccc c ccc}
    \toprule
    & \multicolumn{4}{c}{Archival airborne} & & \multicolumn{3}{c}{UAV-borne (WHU-Hi)} \\
    \cmidrule(lr){2-5} \cmidrule(lr){7-9}
    Strategy & Indian Pines & Salinas & KSC & Pavia U. & & LongKou & HanChuan & HongHu \\
    \midrule
    Peak uncertainty      & $61 \pm 4$ & $151 \pm 19$ & $179 \pm 17$ & $208 \pm 40$ & & $169 \pm 22$ & $217 \pm 48$ & $120 \pm 18$ \\
    Greedy coverage       & $54 \pm 6$ & $66 \pm 2$ & $185 \pm 9$ & $36 \pm 10$ & & $66 \pm 19$ & $160 \pm 28$ & $330 \pm 62$ \\
    Unc.-wt.\ dispersion  & $60 \pm 3$ & $102 \pm 7$ & $185 \pm 11$ & $60 \pm 5$ & & $151 \pm 2$ & $294 \pm 27$ & $369 \pm 33$ \\
    Unc.-ranked NMS       & $60 \pm 3$ & $102 \pm 8$ & $184 \pm 12$ & $60 \pm 5$ & & $151 \pm 2$ & $294 \pm 27$ & $369 \pm 33$ \\
    Stratified unc.       & $68 \pm 1$ & $193 \pm 3$ & $243 \pm 1$ & $265 \pm 2$ & & $241 \pm 6$ & $444 \pm 7$ & $403 \pm 4$ \\
    \midrule
    Lawnmower, scene-wide & $46 \pm 0$ & $78 \pm 0$ & $80 \pm 0$ & $80 \pm 0$ & & $80 \pm 0$ & $80 \pm 0$ & $80 \pm 0$ \\
    Random, scene-wide    & $76 \pm 1$ & $198 \pm 3$ & $289 \pm 2$ & $252 \pm 2$ & & $247 \pm 5$ & $430 \pm 6$ & $377 \pm 5$ \\
    \midrule
    Random, labelled      & $68 \pm 1$ & $172 \pm 1$ & $232 \pm 3$ & $240 \pm 3$ & & $249 \pm 5$ & $430 \pm 8$ & $367 \pm 5$ \\
    Lawnmower, labelled   & $46 \pm 0$ & $49 \pm 0$ & $184 \pm 0$ & $52 \pm 0$ & & $80 \pm 0$ & $95 \pm 0$ & $105 \pm 0$ \\
    Systematic, labelled  & $76 \pm 0$ & $201 \pm 0$ & $222 \pm 0$ & $248 \pm 0$ & & $272 \pm 0$ & $484 \pm 0$ & $414 \pm 0$ \\
    \bottomrule
  \end{tabular}
\end{table*}

Waypoint dispersion (Table~\ref{tab:dispersion}) is descriptive: on every scene
the most dispersed policy is a spatial sampling design (random, systematic or
stratified), and none of them covers the most errors on any scene. Among the uncertainty-driven policies, dispersion
spreads more widely than greedy coverage on every scene but KSC, where the two are
level ($185$\,px), for example $151$ against $66$\,px on LongKou and $60$ against
$36$\,px on Pavia University, consistent with its overlap penalty. Peak
uncertainty follows isolated maxima wherever they fall, so its spread varies
widely between seeds (for example $217 \pm 48$\,px on HanChuan), and spread alone
does not yield coverage.

\subsection{Labelled-Region Dependence and Unmasked Evaluation}
\label{sec:roi_control}

In a deployment the whole scene is classified and no pixel is labelled, so the
surface would be computed over all first-stage predictions. To simulate this we
recompute disagreement over the full scene without the labelled mask, normalize
it over the full scene, and let every policy, baselines included, select
anywhere. We measure uncertainty covered over the full scene and, as before,
error coverage on the labelled pixels (Table~\ref{tab:unmasked}).

Dispersion still covers the most uncertainty on all seven scenes, with smaller
values because the mass now spreads over the whole scene. Error coverage, however, changes sharply. On the three UAV-borne scenes
dispersion still covers the most errors ($18.9$, $5.2$ and $3.7\%$, against at
most $1.4$, $1.0$ and $0.7\%$ for the baselines). On Salinas and KSC it covers
none, on Pavia University $2.7\%$, and on Indian Pines the lawnmower covers more
($30.7$ against $27.2\%$). Without the mask, disagreement is
highest on unlabelled pixels; the planner sends its sites there, and those
inspections cannot be scored because no ground truth exists. The unlabelled
pixels are a mixture: unlabelled parts of the trained classes, mixed pixels along
field and road edges, and land cover outside the label set. The benchmarks do not
say which is which, so we cannot tell how much of this disagreement is in-class
ambiguity, which an inspection would resolve like any other error, and how much
is novelty. The error-coverage results of Section~\ref{sec:error_reduction}
therefore hold where the surface is restricted to labelled land cover. Inspecting
unlabelled ground may well be useful in practice, since the map is unreliable
there, but its value cannot be measured on these benchmarks.

\paragraph{A label-free substitute for the mask.}
A deployment has no labels outside the training set, but it does have the
training set. We therefore score every pixel by its mean distance to the $10$
nearest training spectra in the $30$-component feature space (at most $20{,}000$
reference spectra), set the threshold at the $q$-quantile of the same score on
labelled pixels that no member trained on, drop the pixels above it as novel, and
let every policy select on the unmasked surface renormalized over the rest
(Table~\ref{tab:novelty}). The mask recovers part of the gap where the unmasked
surface failed: at $q=0.90$ dispersion covers $42.9$ instead of $27.2\%$ of the
errors on Indian Pines, $12.4$ instead of $2.7\%$ on Pavia University, $10.0$
instead of $0\%$ on KSC and $23.4$ instead of $18.9\%$ on LongKou, against $53.0$,
$31.8$, $56.7$ and $27.2\%$ with the label mask. On Salinas it recovers nothing:
the unlabelled pixels where disagreement is highest are spectrally close to the
training data, so the mask keeps them and the sites stay off the labelled fields.
On HanChuan and HongHu it removes few unlabelled pixels and changes little (on
HanChuan it lowers coverage from $5.2$ to $4.1\%$ at $q=0.90$); the unmasked surface
already does nearly as well as the label mask there. Spectral novelty therefore explains part
of the failure of the unmasked surface but not all of it; the rest is
disagreement on spectra the classifier knows, outside the labelled fields, and a
deployment would need a further, spatial criterion, or labels for the land cover
it wants verified, to steer sites away from it.

\begin{table}[t]
  \caption{Label-free novelty mask: error coverage (\%) of the dispersion policy
  when the surface is restricted by the labelled region (as in the main results),
  by nothing (Table~\ref{tab:unmasked}), or by a novelty mask that drops pixels
  whose spectra are far from the training spectra (threshold at the $q$-quantile
  of in-distribution scores). ``Unlab.\ kept'' is the share of unlabelled pixels
  the mask keeps at $q=0.90$; ``On labels'' the share of inspected pixels that are
  labelled at $q=0.90$. Mean over seeds.}
   \vspace{-0.3cm}
  \label{tab:novelty}
  \small
  \setlength{\tabcolsep}{3pt}
  \begin{tabular}{l ccccc cc}
    \toprule
    Scene & \makecell{Label\\mask} & \makecell{No\\mask} & \makecell{Novelty\\$q=0.90$} & \makecell{Novelty\\$q=0.95$} & \makecell{Novelty\\$q=0.99$} & \makecell{Unlab.\\kept} & \makecell{On\\labels} \\
    \midrule
    Indian Pines & $53.0$ & $27.2$ & $42.9$ & $41.8$ & $31.2$ & $63\%$ & $63\%$ \\
    Salinas      & $45.7$ & $0.0$  & $0.0$  & $0.0$  & $0.0$  & $57\%$ & $0\%$ \\
    KSC          & $56.7$ & $0.0$  & $10.0$ & $8.4$  & $4.3$  & $64\%$ & $7\%$ \\
    Pavia U.     & $31.8$ & $2.7$  & $12.4$ & $8.6$  & $6.3$  & $79\%$ & $31\%$ \\
    LongKou      & $27.2$ & $18.9$ & $23.4$ & $22.6$ & $21.7$ & $61\%$ & $77\%$ \\
    HanChuan     & $6.6$  & $5.2$  & $4.1$  & $5.0$  & $5.3$  & $94\%$ & $54\%$ \\
    HongHu       & $4.0$  & $3.7$  & $3.7$  & $3.8$  & $3.8$  & $94\%$ & $93\%$ \\
    \bottomrule
  \end{tabular}
\end{table}

\begin{table}[t]
  \caption{Unmasked control: the uncertainty surface is computed over the whole
  scene and every policy may select anywhere. Uncertainty covered over the whole
  scene (\%) and error coverage on labelled pixels (\%), mean over seeds (Indian
  Pines: 10, others: 5).}
   \vspace{-0.3cm}
  \label{tab:unmasked}
  \small
  \setlength{\tabcolsep}{2.5pt}
  \begin{tabular}{l ccccccc}
    \toprule
    & Indian P. & Salinas & KSC & Pavia U. & LongKou & HanChuan & HongHu \\
    \midrule
    \multicolumn{8}{l}{\emph{Uncertainty covered}} \\
    Peak uncertainty & $16.0$ & $3.5$ & $1.3$ & $2.2$ & $6.2$ & $1.6$ & $1.5$ \\
    Greedy coverage  & $20.8$ & $7.4$ & $2.0$ & $3.3$ & $9.6$ & $2.5$ & $2.6$ \\
    Dispersion       & $\mathbf{26.2}$ & $\mathbf{10.6}$ & $\mathbf{2.9}$ & $\mathbf{3.9}$ & $\mathbf{12.4}$ & $\mathbf{3.4}$ & $\mathbf{3.1}$ \\
    Lawnmower        & $22.2$ & $3.4$ & $1.1$ & $1.4$ & $1.2$ & $0.3$ & $0.5$ \\
    Random           & $14.7$ & $2.8$ & $1.1$ & $1.6$ & $1.5$ & $0.9$ & $0.7$ \\
    \midrule
    \multicolumn{8}{l}{\emph{Error coverage}} \\
    Peak uncertainty & $21.5$ & $\mathbf{3.8}$ & $0.0$ & $0.6$ & $2.8$ & $0.9$ & $0.8$ \\
    Greedy coverage  & $23.0$ & $0.0$ & $0.0$ & $1.6$ & $15.6$ & $3.8$ & $3.1$ \\
    Dispersion       & $27.2$ & $0.0$ & $0.0$ & $\mathbf{2.7}$ & $\mathbf{18.9}$ & $\mathbf{5.2}$ & $\mathbf{3.7}$ \\
    Lawnmower        & $\mathbf{30.7}$ & $1.7$ & $0.0$ & $0.5$ & $0.5$ & $0.3$ & $0.2$ \\
    Random           & $14.8$ & $3.0$ & $\mathbf{1.0}$ & $1.7$ & $1.4$ & $1.0$ & $0.7$ \\
    \bottomrule
  \end{tabular}
\end{table}

\subsection{Flight-Cost Analysis}
\label{sec:cost}

\begin{figure}[t]
  \centering
  \includegraphics[width=0.6\linewidth]{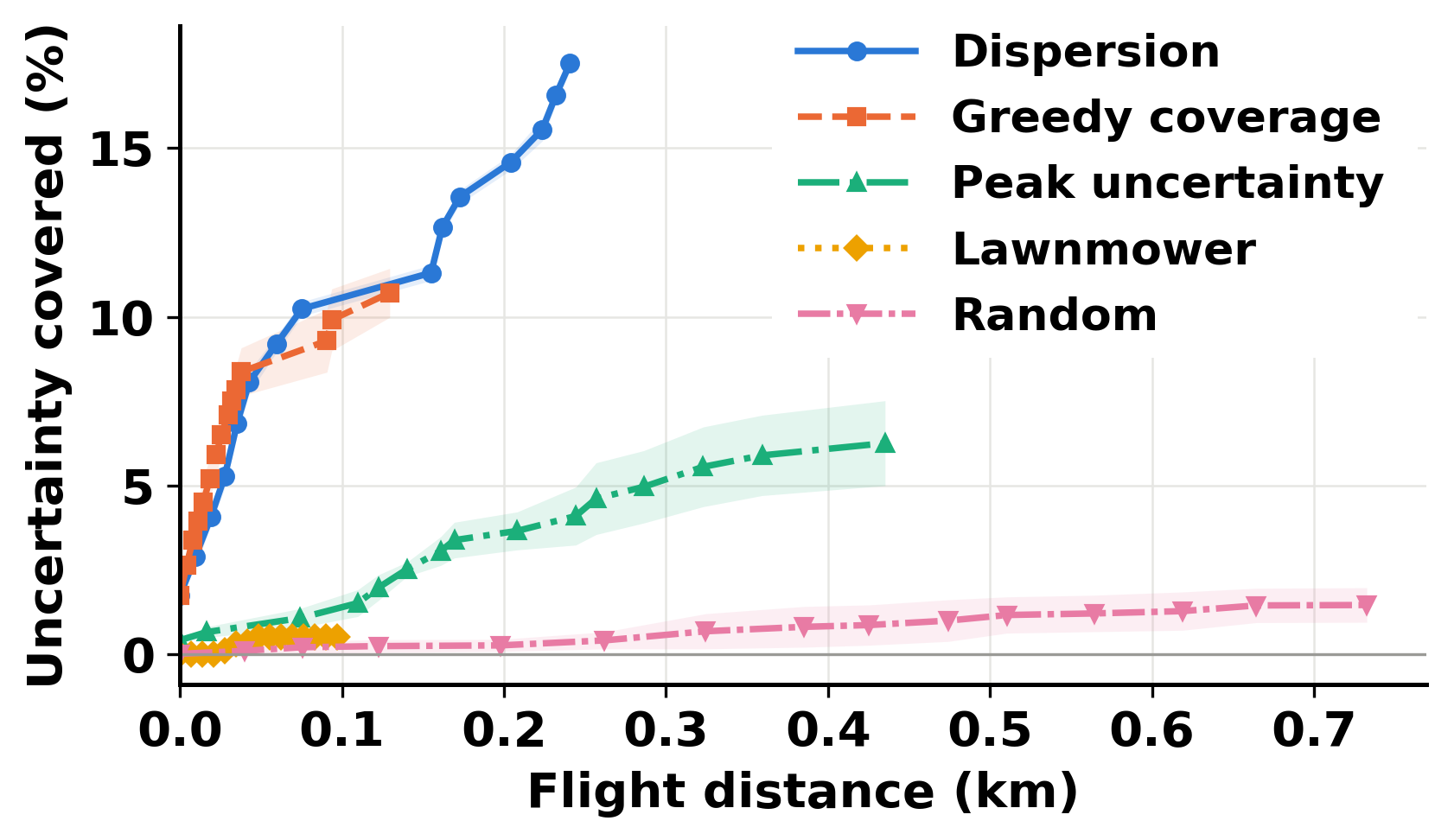}
   \vspace{-0.30cm}
  \caption{Uncertainty covered by the five policies versus cumulative flight distance along the
  nearest-neighbour and 2-opt route on the LongKou UAV scene, mean over five
  seeds with $\pm$ standard-deviation bands.}
  \Description{Line plot of percent uncertainty covered against cumulative
  flight distance in kilometres for five policies, with shaded standard
  deviation bands.}
  \label{fig:cost}
\end{figure}

The metrics above count \emph{waypoints}, treating each inspection as equally
costly. A UAV is constrained by flight distance and energy, so we also measure the
travel each policy's sites imply on the LongKou UAV scene, whose
$\sim$0.46\,m/pixel resolution makes a 15-site tour a realistic close-range
sortie. We report two route lengths: the sites flown in the order the policy
selected them, and the same sites ordered by a nearest-neighbour tour from the
first site refined by 2-opt. Table~\ref{tab:cost} reports means $\pm$ standard
deviations over five seeds, and Figure~\ref{fig:cost} shows uncertainty covered
against cumulative distance along the 2-opt route.

\begin{table}[t]
  \caption{Flight-cost analysis on the LongKou UAV scene (15 sites, 5 seeds, mean
  $\pm$ std, $0.463$\,m/pixel). Path lengths are for the sites flown in selection
  order and after nearest-neighbour and 2-opt ordering. ``Covered'' is the share of
  the uncertainty mass inside the footprints; coverage per km uses the 2-opt
  route.}
   \vspace{-0.3cm}
  \label{tab:cost}
  \small
  \setlength{\tabcolsep}{3pt}
  \begin{tabular}{lcccc}
    \toprule
    Strategy & \makecell{Covered\\(\%)} & \makecell{Path, selection\\order (km)} & \makecell{Path,\\2-opt (km)} & \makecell{Covered\\per km} \\
    \midrule
    Peak uncertainty      & $6.3 \pm 1.3$ & $0.85 \pm 0.04$ & $0.44 \pm 0.05$ & $14.7 \pm 4.1$ \\
    Greedy coverage       & $10.7 \pm 0.7$ & $0.38 \pm 0.10$ & $0.13 \pm 0.00$ & $\mathbf{82.7 \pm 3.7}$ \\
    Unc.-wt.\ dispersion  & $\mathbf{17.5 \pm 0.2}$ & $0.88 \pm 0.19$ & $0.24 \pm 0.01$ & $72.8 \pm 2.8$ \\
    \midrule
    Lawnmower, scene-wide & $0.5 \pm 0.0$ & $0.10 \pm 0.00$ & $0.10 \pm 0.00$ & $5.3 \pm 0.3$ \\
    Random, scene-wide    & $1.5 \pm 0.5$ & $1.54 \pm 0.14$ & $0.73 \pm 0.08$ & $2.0 \pm 0.8$ \\
    \bottomrule
  \end{tabular}
\end{table}

Routing matters: 2-opt shortens every uncertainty-driven
policy's route by $48$ to $73\%$, and after routing a complete 15-site inspection
of LongKou needs $0.13$ to $0.44$\,km, a short distance for a small multirotor, although flight feasibility also depends on speed, hover time and battery reserve, which we do not model. Dispersion covers the most uncertainty ($17.5\%$) over a
$0.24$\,km route. Greedy coverage has the highest coverage per kilometre because
it flies the shortest route ($0.13$\,km), returning to partly inspected areas
(Section~\ref{sec:policies}); dispersion covers $64\%$ more uncertainty for
$0.11$\,km more flight. Peak uncertainty flies the longest route of the three for
the least coverage. The scene-wide baselines cover almost nothing ($0.5$ and
$1.5\%$); the lawnmower's short path runs mostly through unlabelled ground.

\subsection{Sensitivity Analyses}
\label{sec:tau}

\begin{figure}[t]
  \centering
  \includegraphics[width=0.76\linewidth]{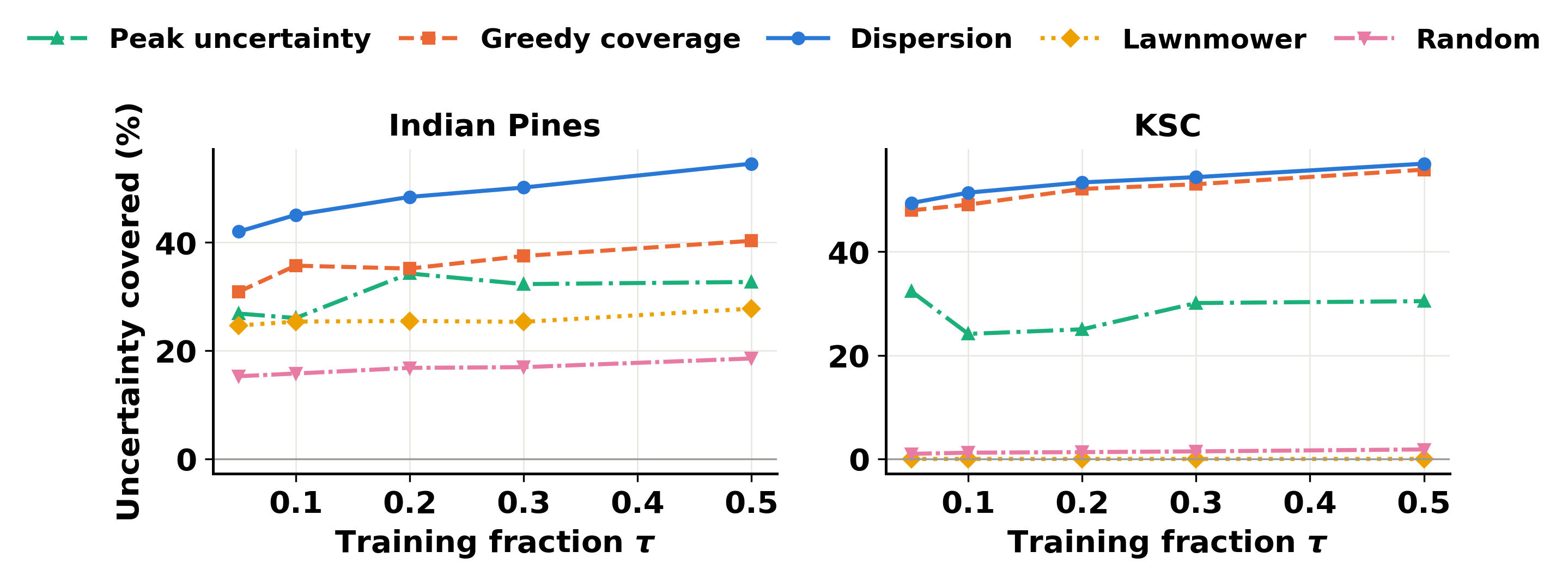}
   \vspace{-0.3cm}
  \caption{Single-seed sensitivity of uncertainty covered to the training fraction $\tau$ on Indian Pines and KSC (seed 0). Across
  $\tau \in \{0.05, 0.1, 0.2, 0.3, 0.5\}$ the dispersion policy remains the
  strongest on both scenes and all three uncertainty-driven policies stay
  above the baselines. Absolute magnitudes are not comparable across $\tau$
  because the surface is renormalized at each setting.}
  \Description{Two line plots of percent uncertainty covered versus the
  training fraction tau for Indian Pines and KSC; the dispersion policy is
  highest at every tau and all three targeted policies stay above the
  baselines.}
  \label{fig:tau}
\end{figure}

\paragraph{Training fraction.}
Beyond the budget, the pipeline's principal hyperparameter is the training
fraction $\tau$. Figure~\ref{fig:tau} sweeps $\tau \in \{0.05, 0.1, 0.2, 0.3,
0.5\}$ on two scenes. Dispersion is the strongest policy at every $\tau$ (Indian
Pines $42.0$ to $54.6\%$, KSC $49.4$ to $57.0\%$), with greedy coverage second at
every $\tau$. All three uncertainty-driven policies stay above the baselines,
peak uncertainty only narrowly on Indian Pines ($26.0$ against $25.4\%$ for the
lawnmower at $\tau = 0.1$). Absolute magnitudes are not
comparable across $\tau$, since the surface is renormalized at each setting.

\paragraph{FOV radius and stride.}
Sweeping $r \in \{5, 7, 11\}$ with stride proportional to $r$, dispersion remains
the best uncertainty-driven policy at every setting on both Indian Pines and
Salinas, with coverage scaling with footprint area (dispersion $29.4$, $45.4$ and
$73.0\%$ on Indian Pines, $19.4$, $30.5$ and $50.4\%$ on Salinas). At the largest
radius on Indian Pines the lawnmower narrows the gap ($61.8$ against $73.0\%$),
since fifteen $23 \times 23$ footprints cover much of a $145 \times 145$ scene;
uncertainty guidance matters in the budget-constrained regime where footprints
are small relative to the scene.

\paragraph{Inspection budget.}
Extending the sweep of Figure~\ref{fig:reduction_curves} from $K{=}5$ to $K{=}30$
preserves the dispersion-first ordering at every budget on both Indian Pines and
Salinas. Peak uncertainty gains slowly on Salinas ($8.0\%$ at $K{=}15$, $14.9\%$ at
$K{=}30$), against $30.5$ and $48.8\%$ for dispersion.

\paragraph{Ensemble size and composition.}
On Indian Pines (seeds 0 to 2, main-protocol AUC), the error-ranking AUC rises
with $M$ with diminishing returns ($0.751$, $0.791$, $0.813$ and $0.822$ for $M =
3, 5, 7, 9$), while the uncertainty dispersion covers stays between $44.6$ and
$46.2\%$. An all-random-forest ensemble matches the mixed ensemble on AUC ($0.787$
against $0.791$) and lets dispersion cover more uncertainty ($49.5$ against
$45.8\%$).

\paragraph{Candidate-grid stride.}
On Indian Pines and Salinas, refining the candidate grid from stride $5$ to $3$
or coarsening it to $7$ changes the uncertainty dispersion covers by at most $1.2$
points ($46.6$, $45.4$ and $44.3\%$ on Indian Pines; $31.1$, $30.5$ and $29.3\%$ on
Salinas) and its error coverage by at most $3.7$ points, and greedy coverage stays
below dispersion at every stride.

\paragraph{Tie-breaking of peak uncertainty.}
Peak uncertainty breaks ties among maxima in scan order. Breaking them uniformly at
random instead changes its error coverage by $-0.9$ to $+4.9$ points (largest on
KSC, $27.4$ to $32.3\%$), leaves it far below dispersion on every scene, and
leaves its connected-anomaly wins on Salinas ($26.1$ against $14.0\%$) and Pavia
University ($21.4$ against $7.3\%$) in place. Those wins are not an artifact of
scan order.

\paragraph{Connected-component definition.}
Over all seeds, switching from 4- to 8-connectivity leaves the per-scene winner of
Table~\ref{tab:coverage} unchanged at the $0.7$ threshold on six scenes; on
HongHu peak and dispersion tie under 8-connectivity ($1.4\%$ each). Changing the
footprint radius to $r = 5$ or $r = 11$ (with stride proportional to $r$) also
leaves the winner unchanged on every scene.

\paragraph{Training-subset sampling.}
Stratified rather than uniform $\tau$-sampling changes the AUC by $0.005$ and
the uncertainty dispersion covers by $0.1$ points ($45.7$ against $45.8\%$).
Training all members on one shared $\tau$-fraction instead of independent ones
lowers the main-protocol AUC from $0.791$ to $0.694$ on Indian Pines (seeds 0 to
2) while the uncertainty covered barely changes ($45.2$ against $45.8\%$); on held-out pixels the shared subset is
$0.02$ to $0.07$ lower on all seven scenes. Diversity in the training sample, not
only in the model, is what makes disagreement informative about error. The PCA dimension $d$ and the per-member tree counts
and depths are applied identically across all scenes and policies.

\paragraph{Connected-anomaly threshold.}
The $0.7$ threshold defines the metric, not any policy, so it cannot bias the
comparison toward one planner by construction, but it does decide which
components count. Because $u$ is a multiple of $0.1$ and never lies between $0.4$
and $0.6$, thresholds from $0.5$ to $0.9$ select four distinct pixel sets, $u \ge
0.6$, $0.7$, $0.8$ and $0.9$ ($0.5$ and $0.6$ select the same pixels). We swept
them over all seeds, with both the equal-weight and the size-weighted metric.
Under size weighting, dispersion wins on every scene at $0.6$; at $0.7$ and above
peak uncertainty wins on Salinas and Pavia University, and at $0.8$ and above on
LongKou as well. Indian Pines, KSC, HanChuan and HongHu go to dispersion at every
threshold. The equal-weight metric agrees except on HanChuan and HongHu at $0.6$,
where peak covers a larger share of the many tiny components. Lower
thresholds admit more of the moderate-uncertainty mass that dispersion follows;
higher ones keep only the pixels on which the members spread over three or more
classes, near the maxima that peak targets. The connected-anomaly winner is
therefore a property of both the scene and the definition of an anomaly, and a
practitioner who cares about the most ambiguous regions should pick the threshold
first.

\subsection{Agnosticism to the Uncertainty Estimator}
\label{sec:estimator}

Because the policies operate on a normalized scalar surface, they should work with
any reasonable uncertainty estimator. To test this we replace the tree ensemble
with three estimators on a small multilayer perceptron over the same
PCA-reduced spectra: Monte-Carlo dropout~\cite{gal2016dropout}, a deep
ensemble~\cite{lakshminarayanan2017deepensembles}, and single-model softmax
entropy. For each we compute the surface, apply the same dispersion policy, and
report the error-ranking AUC of the surface (Section~\ref{sec:error_corr}) and the
share of its uncertainty mass that dispersion covers.

\begin{figure}[t]
  \centering
  \includegraphics[width=0.9\linewidth]{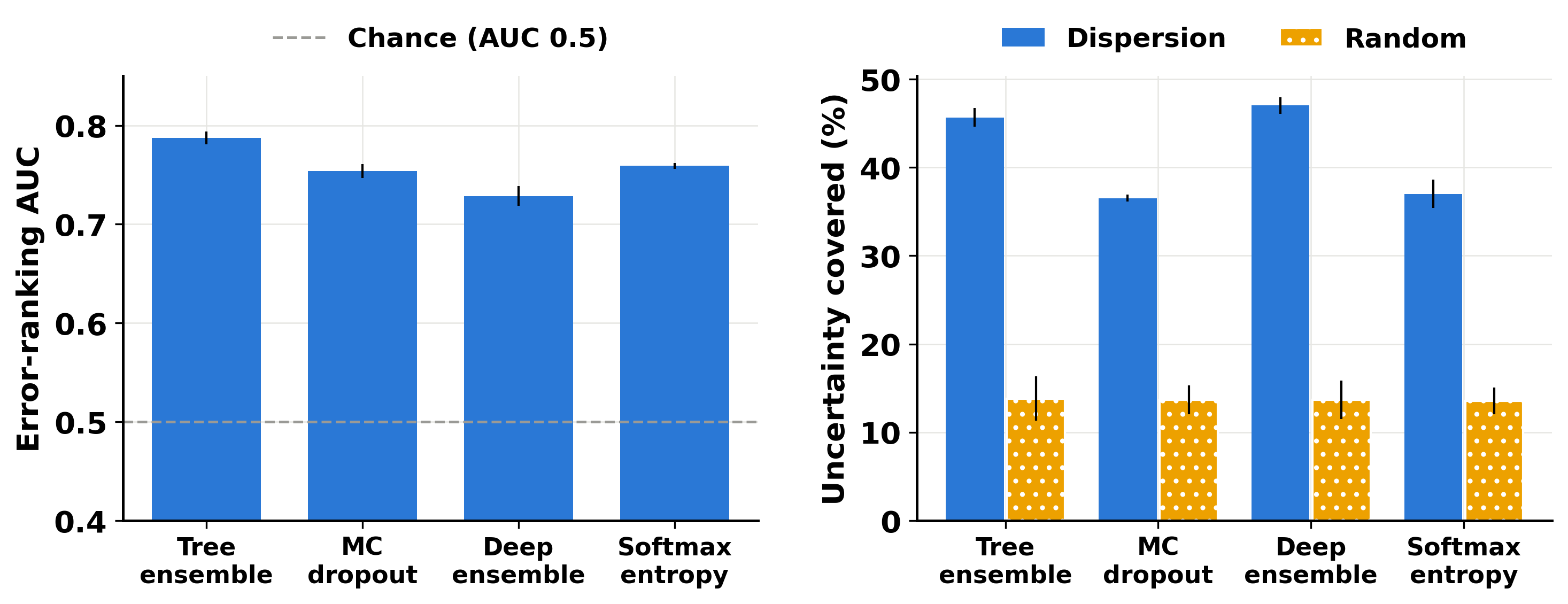}
   \vspace{-0.4cm}
  \caption{\textbf{The planner is agnostic to the uncertainty estimator}
  (Indian Pines). The dispersion policy is applied unchanged to surfaces from
  four estimators, the tree ensemble, MC dropout, a deep ensemble and softmax entropy (mean $\pm$ std over five seeds; Indian Pines only). Left: error-ranking AUC, with
  chance at $0.5$. Right: uncertainty covered by dispersion and by random sites on
  the same surface.}
  \label{fig:estimator}
  \Description{Two bar charts for four uncertainty estimators on Indian Pines: error-ranking AUC of 0.73 to 0.79 above the 0.5 chance line, and uncertainty covered by dispersion (37 to 47 percent) against random sites (about 14 percent).}
\end{figure}

Figure~\ref{fig:estimator} shows the outcome on Indian Pines (five seeds, every
model trained on $10\%$ of the labelled pixels, the deep-ensemble members on the
same subsets as the tree members). All four surfaces rank genuine errors above
chance (main-protocol AUC $0.73$ to $0.79$), and on all four dispersion covers
$37$ to $47\%$ of the uncertainty mass, against $14\%$ for random sites on the
same surfaces. The tree ensemble ranks errors best ($0.788$), ahead of softmax
entropy ($0.759$), MC dropout ($0.754$) and the deep ensemble ($0.729$); dispersion
covers the most on the deep ensemble ($47.0\%$) and tree ensemble ($45.7\%$) and
the least on the two entropy surfaces ($36.5$ to $37.0\%$). On this scene the
CPU-only tree ensemble is thus at least as good as the neural estimators, and a
richer estimator can be substituted without changing the planner. AUC and
uncertainty covered measure different properties of a surface (how well it ranks
errors versus how concentrated its mass is), so they need not move together.

\subsection{A Spectral-Spatial Classifier}
\label{sec:spatialclf}

The pixel-wise ensemble reaches $77\%$ accuracy on Indian Pines, well below
spectral-spatial classifiers. To test whether the planning result depends on a
weak classifier, we train the same members on $50$ spectral-spatial features: the
$30$ principal components plus the local means of the first $10$ components over
$5\times5$ and $11\times11$ windows. We build surfaces under the main protocol and
under the blocked protocol, whose $7$-pixel buffer exceeds the $5$-pixel window
half-width, so no test pixel's window reaches a training block
(Table~\ref{tab:spatialclf}; Indian Pines, Salinas and Pavia University, seeds 0
to 4).

Under the main protocol the spatial features raise accuracy to $92.6$, $99.1$ and
$98.8\%$ and the AUC to $0.86$ to $0.88$, and dispersion covers $55$, $81$ and
$62\%$ of the now much rarer errors, $2.0$ to $13.5$ times as many as the best
region-matched baseline. These figures overstate the classifier, since the window
around an evaluated pixel overlaps the training pixels of every member. Under the
blocked protocol the gains shrink: accuracy is $40.3$, $87.5$ and
$87.5\%$ against $34.1$, $80.9$ and $79.0\%$ for pixel features (Indian Pines is
low for both), and the AUC is $0.645$ to $0.703$ against $0.643$ to $0.652$.
Dispersion still covers $1.2$ to $4.4$ times as many held-out errors as the best
region-matched baseline, close to the pixel-wise ratios ($1.3$ to $3.1$). The
planning advantage therefore does not depend on a weak classifier, and on ground
far from the training data a spectral-spatial classifier of this kind ranks its
own errors little better than a pixel-wise one.

\begin{table}[t]
  \caption{Pixel-wise against spectral-spatial features: accuracy (\%), AUC of
  $u$, error coverage of dispersion (\%) and its ratio to the best region-matched
  baseline, under the main and the blocked protocol ($32$-pixel blocks). Mean
  over seeds 0 to 4.}
   \vspace{-0.3cm}
  \label{tab:spatialclf}
  \small
  \setlength{\tabcolsep}{3pt}
  \begin{tabular}{ll cccc cccc}
    \toprule
    & & \multicolumn{4}{c}{Main protocol} & \multicolumn{4}{c}{Blocked} \\
    \cmidrule(lr){3-6} \cmidrule(lr){7-10}
    Scene & Features & Acc. & AUC & Disp. & Ratio & Acc. & AUC & Disp. & Ratio \\
    \midrule
    Indian Pines & pixel   & $77.2$ & $0.788$ & $54.3$ & $1.8$ & $34.1$ & $0.643$ & $69.6$ & $1.3$ \\
                 & spatial & $92.6$ & $0.869$ & $54.8$ & $2.0$ & $40.3$ & $0.656$ & $69.7$ & $1.2$ \\
    Salinas      & pixel   & $94.2$ & $0.825$ & $45.7$ & $6.0$ & $80.9$ & $0.648$ & $33.0$ & $2.6$ \\
                 & spatial & $99.1$ & $0.859$ & $81.4$ & $6.7$ & $87.5$ & $0.645$ & $31.8$ & $2.5$ \\
    Pavia U.     & pixel   & $91.1$ & $0.827$ & $31.8$ & $6.6$ & $79.0$ & $0.652$ & $34.0$ & $3.1$ \\
                 & spatial & $98.8$ & $0.882$ & $62.0$ & $13.5$ & $87.5$ & $0.703$ & $46.2$ & $4.4$ \\
    \bottomrule
  \end{tabular}
\end{table}

\subsection{Cost-Aware Spatial Coverage}
\label{sec:costaware}

\begin{figure}[t]
  \centering
  \includegraphics[width=0.5\linewidth]{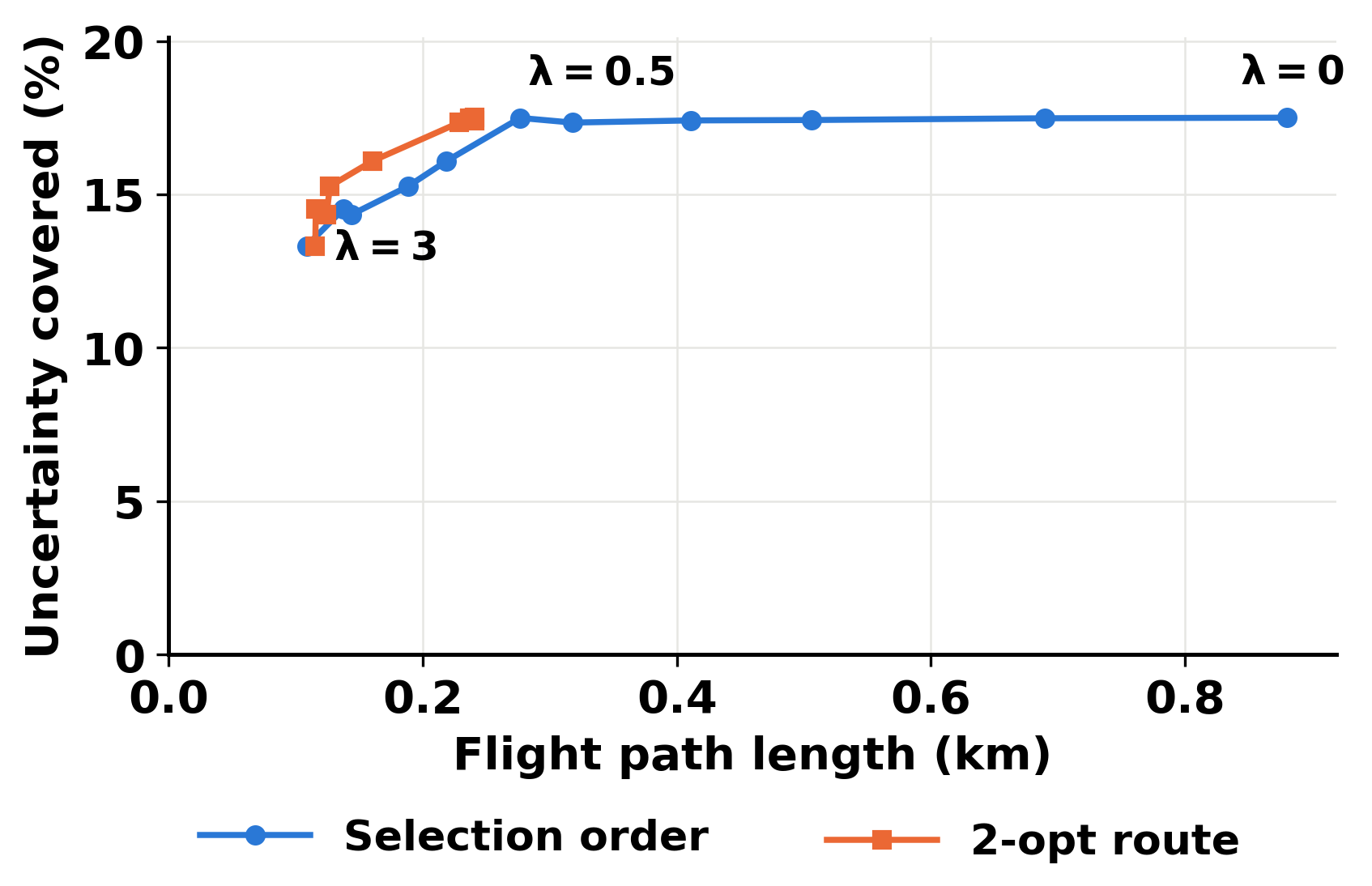}
   \vspace{-0.3cm}
  \caption{\textbf{Cost-aware trade-off on LongKou.} Uncertainty covered versus
  flight-path length as the travel weight $\lambda$ varies from $0$ (dispersion)
  to $3$ (mean over 5 seeds), with the sites flown in selection order and after
  nearest-neighbour and 2-opt ordering. Labels mark $\lambda$ on the
  selection-order curve. The penalty acts on distance, not energy.}
  \label{fig:pareto}
  \Description{Line plot of percent uncertainty covered against flight path length in kilometres as the travel weight increases from 0 to 3, for sites flown in selection order and after 2-opt ordering.}
\end{figure}

The cost-aware policy~\eqref{eq:costaware} subtracts a travel penalty
$\lambda\, d/\mathrm{diag}$ from the dispersion score, where $d$ is the distance
from the previously selected site; at $\lambda=0$ it is the dispersion policy.
Figure~\ref{fig:pareto} traces the trade-off on LongKou for both route
definitions. Flown in selection order, raising $\lambda$ from $0$ to $0.5$ cuts
the path from $0.88$ to $0.28$\,km (to $31\%$ of its length) at no cost in coverage ($17.5\%$ of
the uncertainty mass at both). That saving, however, comes almost entirely from the order in which the
sites are visited, which a route optimizer recovers anyway: with both routes
ordered by 2-opt, $\lambda = 0.5$ leaves the path at $0.24$\,km. Only larger
weights shorten the routed tour: $\lambda = 0.75$ shortens it by a third, from
$0.24$ to $0.16$\,km, while keeping $92\%$ of the uncertainty covered ($17.5\% \to
16.1\%$), and $\lambda = 1$ halves it ($0.13$\,km) while keeping $87\%$. At
$\lambda = 3$ the routed path is $48\%$ of the original and $76\%$ of the coverage
remains. A single
interpretable weight thus trades a modest share of coverage for a shorter tour,
and the routing step should be applied whatever the weight. The heuristic counts
distance only and does not model energy, so choosing $\lambda$ for a given
battery is left to the operator.

\subsection{Comparison to Gaussian-Process Selection}
\label{sec:gpmi}

Finally, we compare against two Gaussian-process (GP) selection rules from
informative path planning. Both fit a GP with an RBF kernel to the surface on a
random subsample of $500$ pixels, as full-scene GP inference is intractable at
these resolutions, and both choose $K = 15$ sites with the same footprint as our
policies and no travel limit, so the comparison isolates the selection
objective. \emph{GP mutual information}
(GP-MI)~\cite{singh2009informative,krause2008near}, the classical objective,
greedily picks the locations that maximize the mutual information between visited
and unvisited sites. \emph{GP-UCB} is a value-aware variant in the style of batch
Bayesian optimization: the GP is fit to the footprint mean of $U$, and the planner
repeatedly takes the non-overlapping candidate footprint with the largest upper
confidence bound $\mu + 2\sigma$, adding it to the GP with its predicted value so
that nearby uncertainty shrinks. Travel-budgeted formulations such as submodular
orienteering change the problem rather than the objective, and are outside the
scope of this paper.

\begin{figure}[t]
  \centering
  \includegraphics[width=0.8\linewidth]{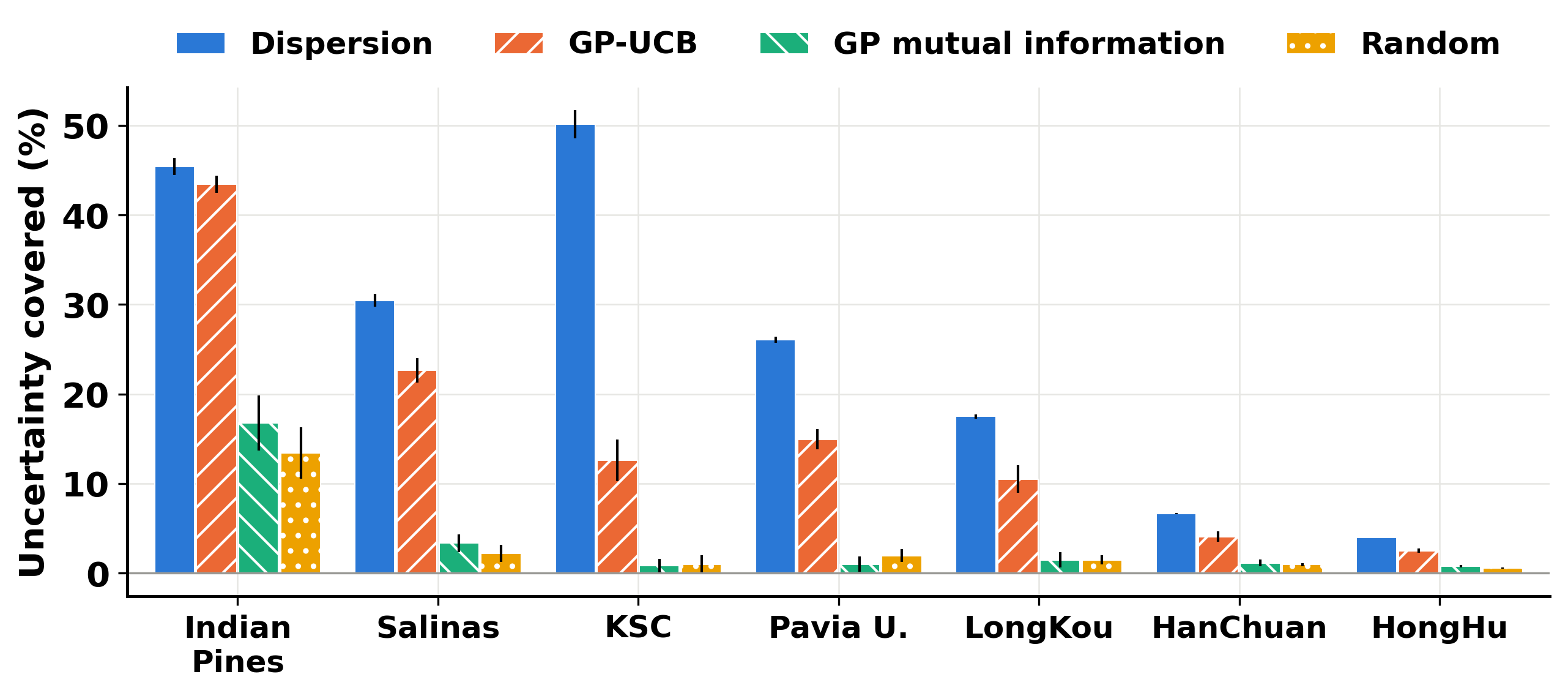}
   \vspace{-0.30cm}
  \caption{\textbf{Comparison to Gaussian-process selection.}
  Uncertainty covered by the dispersion policy (ours), GP mutual information~\cite{singh2009informative}, GP-UCB and random sites, each with 15 sites and no travel limit, on all seven scenes (mean $\pm$ std over seeds).}
  \label{fig:gp}
  \Description{Grouped bar chart of percent uncertainty covered on seven scenes by dispersion, GP-UCB, GP mutual information and random sites; dispersion is highest on every scene.}
\end{figure}

Table~\ref{tab:classical} and Figure~\ref{fig:gp} report the comparison. GP-MI
covers about as much uncertainty as random sites, within $3.4$ points on every
scene and less on KSC and Pavia University, against $4.0$ to $50.1\%$ for
dispersion. This is a mismatch of
objectives, not a weak implementation of the same one: once the kernel is fit,
the mutual information of a site depends only on where it lies relative to the
other sites, not on the uncertainty values there, so GP-MI spreads its sites
evenly to learn the field everywhere. GP-UCB uses the values and does much
better, close to dispersion on Indian Pines ($43.4$ against $45.4\%$), but it
falls behind on every larger scene, to $12.6$ against $50.1\%$ on KSC and
$58$ to $74\%$ of dispersion's coverage on the other five. A smooth GP fit to a few
hundred samples cannot resolve a surface that is zero on most pixels and high on
small, scattered fragments; on KSC, where only $2\%$ of the scene is labelled,
almost every sample falls on zero. Dispersion reads the surface directly, which
on a surface that is already fully known is both cheaper and more accurate. The
comparison shows that the informative-path-planning objectives should not be
adopted unchanged for inspecting a known uncertainty map, not that our planner is
a better solver of them.

\begin{table}[t]
  \caption{Comparison to Gaussian-process selection: uncertainty covered (\%),
  mean $\pm$ std over seeds, $K=15$ sites with no travel limit for every policy.}
   \vspace{-0.3cm}
  \label{tab:classical}
  \small
  \begin{tabular}{l cccc}
    \toprule
    Scene & \makecell{Dispersion\\(ours)} & GP-MI & GP-UCB & Random \\
    \midrule
    Indian Pines & $\mathbf{45.4 \pm 0.9}$ & $16.8 \pm 3.1$ & $43.4 \pm 1.0$ & $13.4 \pm 2.9$ \\
    Salinas      & $\mathbf{30.5 \pm 0.7}$ & $3.3 \pm 1.0$  & $22.6 \pm 1.4$ & $2.2 \pm 1.0$ \\
    KSC          & $\mathbf{50.1 \pm 1.5}$ & $0.8 \pm 0.8$  & $12.6 \pm 2.3$ & $1.0 \pm 1.0$ \\
    Pavia U.     & $\mathbf{26.0 \pm 0.3}$ & $1.0 \pm 0.9$  & $15.0 \pm 1.1$ & $1.9 \pm 0.7$ \\
    LongKou      & $\mathbf{17.5 \pm 0.2}$ & $1.5 \pm 0.8$  & $10.5 \pm 1.5$ & $1.5 \pm 0.5$ \\
    HanChuan     & $\mathbf{6.6 \pm 0.1}$  & $1.1 \pm 0.4$  & $4.1 \pm 0.6$  & $0.9 \pm 0.2$ \\
    HongHu       & $\mathbf{4.0 \pm 0.0}$  & $0.8 \pm 0.1$  & $2.5 \pm 0.2$  & $0.6 \pm 0.1$ \\
    \bottomrule
  \end{tabular}
\end{table}

\section{Discussion}
\label{sec:limitations}

\subsection{Implications for Spatial Analytics and Scientific Verification}
\label{sec:implications}

The results treat classifier uncertainty as a spatial object rather than a
per-pixel scalar. A bare uncertainty value says how ambiguous a pixel is; its
\emph{spatial arrangement}, whether ambiguity is scattered across many small
components or concentrated in a few, and how the sensor footprint tiles that
arrangement, determines how a limited verification budget should be allocated.
Exploiting the arrangement is what lets a fixed budget cover several times more
genuine errors than uncertainty-agnostic sampling of the same labelled region,
provided the surface is restricted to labelled land cover.
The best policy for covering uncertainty mass and errors is the classical
coverage greedy on every scene, and a non-overlapping selection of high-mass
footprints matches it. The design choice that matters is therefore the unit of
scoring: whole footprints by the uncertainty mass they add, rather than single
pixels (peak uncertainty), footprint means (greedy coverage) or a fixed spatial
quota (stratified sampling). The exact marginal-gain rule adds a guarantee rather
than measured coverage. This is consistent with Section~\ref{sec:morphology}:
almost no high-uncertainty component is larger than a footprint, so the best
footprints seldom compete for the same pixels. The best policy for
covering connected anomalies, by contrast, depends on the scene and on the
anomaly threshold in a way no single scalar descriptor captures.

\subsection{Limitations and Future Work}
\label{sec:limits}

\emph{Uncertainty estimate.} The surface is the disagreement of a five-member
classifier ensemble, each member trained on its own random subset of the labelled
pixels. It does not
separate epistemic from aleatoric uncertainty, depends on ensemble diversity and
on $\tau$, and takes only a few distinct values, so many pixels tie. The main
error-ranking protocol lets some evaluated pixels appear in some members'
training sets, which inflates its AUC; the held-out protocol of
Table~\ref{tab:auc} removes this, and we report it as the primary measure. On
spatially blocked test ground the archival scenes' AUC drops to $0.63$ to $0.69$
with $32$-pixel blocks and to $0.56$ to $0.62$ with $128$-pixel blocks, so a
surface built from scattered labels transfers less well to fields the classifier
has not seen nearby. The planning advantage survives at every block size we tried
but shrinks as blocks grow (Section~\ref{sec:blocked_eval}), and the smallest
blocks are only a few metres wide on the UAV-borne scenes. Spectral-spatial
features improved the blocked AUC by at most $0.05$ (Section~\ref{sec:spatialclf}); a surface
that transfers to distant ground would need training designed for it.

\emph{Labelled region.} The surface is masked to labelled pixels and errors are
counted there. The region-matched baselines (Section~\ref{sec:protocol}) separate
the value of the uncertainty surface from knowledge of where labels are. Without
the mask (Section~\ref{sec:roi_control}), dispersion still covers the most
uncertainty, but on Salinas and KSC it directs its sites to unlabelled ground,
where error coverage cannot be measured. A label-free mask that drops spectrally
novel pixels recovers part of the error coverage on four scenes but none on
Salinas, so a deployment would still need a way to separate in-class ambiguity
outside the labelled fields from ground it does not want verified.

\emph{Footprint and platform.} The footprint is fixed in pixels, so its physical
size spans $0.6$\,m to $300$\,m across scenes and the archival scenes do not
correspond to a low-altitude platform. A constant-physical-footprint analysis
would sharpen the comparison between archival and UAV-borne scenes.

\emph{Planning.} The planner selects a fixed number of sites. Dispersion carries
the classical $(1-1/e)$ guarantee for this cardinality-constrained objective, but
the peak, greedy-coverage and cost-aware policies are heuristics, and none of the
policies optimizes a route. We do not model flight dynamics, energy use or a
flight budget; under a tour or energy budget the problem becomes submodular
orienteering~\cite{singh2009informative,zhang2016submodular}, which we do not
address. The method also assumes
the scene has been surveyed and does not address online planning in a partially
observed scene, where a spatial model would have to predict uncertainty into
unobserved regions, as in Gaussian-process informative path
planning~\cite{singh2009informative,popovic2020ipp}.

\emph{From coverage to correction.} We measure whether inspected footprints
contain errors, not whether inspection corrects them. Measuring classifier
improvement after inspected labels are corrected would close the loop. Because
the planner consumes only a scalar surface, transfer to RGB or thermal imagery
reduces to producing a surface whose disagreement still tracks error, validated
by the protocol of Section~\ref{sec:error_corr}.

\section{Conclusion}
\label{sec:conclusion}
We studied where a budget-limited follow-up inspection should be spent after a
hyperspectral survey has been classified, as a spatial-analytics problem: the
errors a classifier makes are spatially structured, so the arrangement of
classifier uncertainty determines where physical verification pays off. With a
per-pixel uncertainty surface from ensemble disagreement and greedy,
field-of-view-aware coverage of the uncertainty mass, we showed on seven
benchmark scenes that disagreement ranks errors on pixels no ensemble member has
seen with an AUC of $0.74$ to $0.82$, and that the coverage greedy covers up to
$57\%$ of genuine classifier errors with 15 sites, $1.7$ to $18$ times as many as
uncertainty-agnostic sampling of the same labelled region. A non-overlapping
selection of high-mass footprints does as well and stratified sampling does
worse, so what matters is scoring footprints by uncertainty mass. The advantage
holds when planning and scoring are confined to spatially held-out blocks of $32$
to $128$ pixels and with a spectral-spatial classifier, although on blocked
ground disagreement ranks errors less well and the advantage shrinks as blocks
grow. These error-coverage gains hold where the surface is restricted to
labelled land cover; over the whole scene, disagreement concentrates on
unlabelled ground, and masking spectrally novel pixels recovers only part of the
loss. We further found that the best policy for covering connected
high-uncertainty regions depends on the scene and on the threshold that defines
them, in a way a scalar concentration measure does not predict; that the
planning layer works with tree ensembles and deep estimators alike; that a single
travel weight shortens the routed tour by a third while keeping $92\%$ of the
uncertainty covered; and that Gaussian-process selection covers less uncertainty
than the coverage greedy on every scene, by mutual information about as little as
random sites. The result is not a new classifier or solver but a
spatial-analytics treatment of a decision, where to verify, that prior work has
largely left implicit.

The framework runs on commodity CPUs. The code, which rebuilds every uncertainty
surface, table and figure in this paper, is available at
\url{https://anonymous.4open.science/r/tsas-uncertainty-inspection-7638}. Natural next steps are an online setting, in
which a spatial model predicts uncertainty into unobserved regions and the plan
adapts as inspection proceeds, and validation against field-collected inspection
outcomes.

\section*{Ethics and Privacy Statement}
This work uses public benchmark hyperspectral datasets and introduces no personal
or sensitive data. The WHU-Hi datasets~\cite{zhong2020whuhi} are used under their
academic-research terms. Uncertainty-guided inspection planning is intended to
make environmental and agricultural monitoring more efficient; as with any
remote-sensing method, deployment over inhabited areas should respect applicable
privacy and airspace regulations.

\bibliographystyle{ACM-Reference-Format}
\bibliography{references.bib}

\end{document}